\documentclass[]{imag-ms-template}
\usepackage{siunitx}
\usepackage{multirow}

\title{Eddeep: a deep-learning framework for fast eddy-current distortion correction in diffusion MRI} 

\author{Antoine Legouhy$^{1,2,\ast}$, Ross Callaghan$^{1,3}$, Yuchuan Qiao$^{4}$, Whitney Stee$^{5,6}$,\\ Philippe Peigneux$^{5,6}$, Hojjat Azadbakht$^{3}$, Hui Zhang$^{1}$\\[6pt]
\small $^{1}$ Hawkes Institute \& Department of Computer Science, University College London, London, UK\\[3pt]
\small $^{2}$ Institut Pasteur, Université Paris Cité, Unité de Neuroanatomie Appliquée et Théorique, France\\
\small $^{3}$ AINOSTICS ltd., Manchester, UK\\
\small $^{4}$ Institute of Science and Technology for Brain-Inspired Intelligence, Fudan University, Shanghai, China\\
\small $^{5}$ UR2NF – Neuropsychology and Functional Neuroimaging Research Unit\\
\small affiliated at CRCN - Centre for Research in Cognition and Neurosciences\\
\small and UNI - ULB Neuroscience Institute, Université Libre de Bruxelles (ULB), Brussels, Belgium\\[3pt]
\small $^{6}$ GIGA - Cyclotron Research Centre - In Vivo Imaging, University of Liège (ULiège), Liège, Belgium\\[4pt]
\small $^{\ast}$Correspondence: alegouhy@pasteur.fr}

\begin{document} 

\maketitle


\begin{abstract}
Diffusion MRI (dMRI) relies on diffusion-weighted echo-planar imaging, which is highly susceptible to eddy-current-induced geometric distortions. These distortions vary across diffusion volumes according to gradient strength and direction, causing between-volume misalignment that can bias downstream microstructural analyses. Current state-of-the-art correction methods, such as FSL Eddy, achieve high-quality correction through iterative prediction-correction schemes but are computationally expensive and require a sufficiently dense sampling of q-space. We propose \emph{Eddeep}, a deep-learning framework for fast eddy-current distortion correction in dMRI. \emph{Eddeep} decomposes the problem into two stages. First, a supervised image translation network standardises the appearance of diffusion-weighted and $b=0$ images, removing contrast differences that hinder reliable registration. Second, an unsupervised registration network estimates both eddy-current distortion and between-volume head motion parameters under a physics-inspired quadratic distortion model, enabling correction in a single forward pass. The method was trained on UK Biobank data and evaluated on both in-domain (UK Biobank) and out-of-domain (Memodyn) datasets. Across a range of complementary metrics, including between-volume jitter, diffusion kurtosis imaging residuals, signal irregularity, and mutual information, \emph{Eddeep} achieved correction quality comparable to that of FSL Eddy while substantially reducing inference time and without requiring any particular q-space sampling scheme. These results demonstrate that deep learning can provide accurate and efficient eddy-current distortion correction without relying on iterative optimisation, supporting the development of faster diffusion MRI processing pipelines for large-scale studies and clinical deployment.

The code is available at:~\url{https://github.com/CIG-UCL/eddeep}.
\end{abstract}

\section{Introduction}
\label{intro}
In magnetic resonance imaging (MRI), diffusion-weighted imaging (DWI) is a technique designed to indirectly probe the intricate microstructure of biological tissues in vivo, in a non-invasive and non-ionising fashion. The diffusion or random thermal motion of water molecules in biological tissues is hindered and restricted by obstacles such as cell membranes and organelles, resulting in a pattern of diffusion that reflects the local tissue architecture. By applying magnetic field gradients of controlled strength and duration (parameterised by the $b$-value) and direction (the $b$-vector), the MRI signal can be sensitised to this pattern of diffusion. Using an appropriate set of $b$-values and $b$-vectors, the resulting signals can then be used to infer tissue properties at scales far below the imaging resolution.

DWI has applications in both clinical and neuroscience research settings. Clinically, it is central to applications including the early diagnosis of acute ischemic stroke~\citep{fiebach2002}, tumor characterisation~\citep{maier2010}, and preoperative surgical planning~\citep{berman2009}. In neuroscience, diffusion tensor imaging (DTI)~\citep{basser1994} has been widely used to characterise white matter alterations in neurological and psychiatric disorders~\citep{lebihan2012}, while more recent multi-shell acquisitions have enabled biophysical models of tissue microstructure such as NODDI~\citep{Zhang2012,Kamiya2020,kraguljac2023} that address some of the limitations of DTI, moving towards the goal of non-invasive in vivo virtual biopsy~\citep{alexander2019}. These increasingly sophisticated applications demand ever richer acquisitions: while DTI requires as few as six gradient directions at a single $b$-value, modern multi-shell protocols can necessitate over a hundred volumes spanning multiple $b$-values and directions.

To achieve acceptable scan duration, spin-echo single-shot echo-planar imaging (EPI) sequences are used~\citep{turner1990,skare2010}. However, the speed of EPI comes at the cost of a heightened sensitivity to off-resonance effects, which manifest as two distinct families of geometric distortions. The first, susceptibility-induced distortions, arise from spatial inhomogeneities in the static magnetic field caused by differences in magnetic susceptibility between adjacent tissues, most prominently at air-tissue interfaces~\citep{jezzard1999}. These distortions depend on the subject's anatomy and, in the absence of motion, are identical across all volumes of an acquisition. The second, eddy-current-induced distortions, arise from the rapid switching of the diffusion-sensitizing gradients, which produce unwanted magnetic fields that interfere with spatial encoding during readout~\citep{chang1992,jezzard1998}. While susceptibility distortions are common to all EPI sequences, eddy-current distortions are specific to diffusion MRI. Crucially, because the diffusion gradients differ in strength and direction from volume to volume, so do the resulting eddy-current distortions, causing between-volume misalignment that, if left uncorrected, propagates into and biases all downstream modelling and analysis~\citep{haselgrove1996}. In this paper, we specifically address the correction of eddy distortions.

\begin{figure}[!h]
    \centering
    \includegraphics[width=0.8\linewidth]{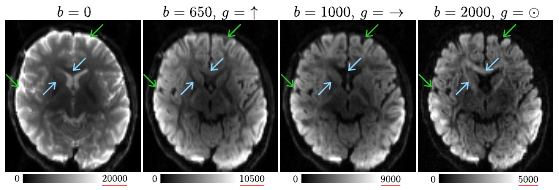}
    \caption{Images are incompatible for registration, their intensity patterns are too different to be compared. Same subject, different $b$-values $b$ and gradient direction $\boldsymbol{g}$. Green arrows: vanishing CSF boundary as $b$ increases, blue arrows: orientation specific attenuations, red underline: signal magnitude decreasing as $b$ increases (despite the same noise level).}
    \label{imgincomp}
\end{figure}

Several strategies exist to mitigate eddy-current effects at the hardware and acquisition level~\citep{bernstein2004}. Modern MRI systems employ actively shielded gradient coils~\citep{mansfield1986} and gradient pre-emphasis~\citep{ahn1991}, which substantially reduce eddy currents induced in surrounding conductive structures. These measures are largely sufficient for conventional imaging sequences. However, diffusion MRI poses a particular challenge because the diffusion-sensitizing gradients are much stronger than typical imaging gradients, inducing proportionally larger eddy currents that may not be fully compensated by hardware solutions alone, especially for high $b$-values ($>$ 2000 s/mm\textsuperscript{2}).
The twice-refocused spin echo sequence~\citep{reese2003} and bipolar diffusion gradient schemes~\citep{alexander1997} can further reduce eddy-current-induced phase errors through gradient waveform design. However this is at the cost of increased echo time and reduced signal-to-noise ratio (SNR), so they are not used in recent large-scale databases such as UK~Biobank~\citep{miller2016} and the Human Connectome Project~\citep{sotiropoulos2013}. As a result, residual distortions of clinically relevant magnitude persist, necessitating correction after data acquisition. 
Post-acquisition correction is typically formulated as an image registration problem.
An initial strategy was to align volumes acquired with non-zero $b$-values, thus eddy-distorted, with the $b=0$ volume which is not. This can work at moderate $b$-values with a suitable similarity metric, but becomes increasingly difficult as the $b$-value increases, typically beyond 1000\,s/mm², because correspondences between non-zero $b$-value and $b=0$ volumes are severely disrupted by volume-specific signal attenuations induced by varying directions and $b$-values. 
In particular, the outer cerebro-spinal fluid (CSF) is a vanishing boundary beyond very low $b$-values; very different attenuation patterns appear in areas of anisotropic diffusion such as white matter; and the signal-to-noise ratio (SNR) decreases as the $b$-value increases (see Fig.~\ref{imgincomp} for an illustration). For the registration process to produce the desired alignment, thus the correction, the two images to be registered must be deemed sufficiently comparable by the similarity metric chosen. Attempts were made using similarity metrics designed to handle large intensity differences, such as correlation ratio~\citep{haselgrove1996,bastin1999} and mutual information~\citep{mangin2002,rohde2004}. But even these do not behave well for high $b$-values~\citep{rohde2004,andersson2016, bastin1999}. 

Given the impracticability of using the $b=0$ image as a registration target, a new prediction-correction strategy has emerged~\citep{andersson2016}. It is an iterative algorithm with each iteration consisting of two steps: first, a prediction step estimates from the data, at the current level of correction, a distortion-free version of each volume; second, a correction step aligns each volume to the distortion-free prediction of itself. By iterating these steps repeatedly, the predictions become progressively more accurate, enabling increasingly more precise distortion correction.
In FSL~Eddy~\citep{andersson2016}, arguably the most popular tool for eddy-current distortion correction, the prediction step uses a Gaussian process~\citep{andersson2015} to estimate the distortion-free version of each volume from those with similar $b$-vectors and the same $b$-value. It is designed for the common multi-shell acquisitions.
The most recent version of Tortoise~\citep{tortoisev4} has also adopted this strategy but the predictions are made by fitting a MAP-MRI~\citep{ozarslan2013} model to the entire dataset at each iteration. It is designed to support arbitrary $b$-value and $b$-vector sampling.
While both tools achieve good correction, the overall processing remains time-consuming, particularly without GPU acceleration. The prediction-correction scheme is iterative and the correction itself is performed using traditional registration for which a new iterative optimisation has to be restarted from scratch for each volume pair. For advanced diffusion MRI, fast preprocessing is not merely an engineering convenience, but a prerequisite for moving from research settings to deployable clinical tools.

Deep-learning methods, enabling transformation parameters to be predicted in a single forward pass, have been developed for fast image registration~\citep{devos2017,voxelmorph1}.
In diffusion MRI, these have already been successfully applied to fast correction of susceptibility-induced distortions~\citep{duong2020,legouhy2022,zaid2024} for blip-up/blip-down acquisitions. However, to the best of our knowledge, no deep-learning approach has yet been proposed for eddy-current distortion correction. Unlike for susceptibility distortion correction where images of similar contrast have to be matched (blip-up and blip-down pair), eddy-current distortion correction must operate across diffusion-weighted volumes whose correspondences are disrupted by the large contrast differences. We hypothesise that this obstacle can be addressed by first bringing the images into closer appearance correspondence through deep-learning-based image translation. Such strategies have been explored already in diffusion MRI, for instance for synthesising $b=0$ images from T1-weighted images~\citep{schilling2019}.

In this paper, we introduce \emph{Eddeep}, a novel approach for fast eddy-current distortion correction in dMRI that leverages these deep-learning advances. Unlike conventional iterative prediction-correction schemes, \emph{Eddeep} addresses contrast standardisation and geometric distortion correction in a single two-stage pass through two models applied in sequence:
\begin{enumerate}
    \item An image translation model to restore correspondences between volumes. Regardless of its $b$-value and gradient direction, each volume is translated to a version with preserved anatomy but a target contrast that does not contain CSF outer boundaries or gradient direction-dependent attenuation patterns. The model is trained in a supervised fashion on previously-corrected data.
    \item A registration model to jointly estimate head motion and eddy-distortion parameters between an uncorrected translated DW volume and a distortion-free translated $b=0$ volume. Constraints inspired by acquisition physics are incorporated into the estimation to ensure plausible eddy-current distortions in EPI. The model is trained in an unsupervised fashion.
\end{enumerate}

\emph{Eddeep} therefore maps each input volume to an undistorted geometry referenced to $b=0$, does not rely on any specific gradient sampling scheme, and performs correction at inference time several times faster than traditional FSL~Eddy on the same processing device. A preliminary version of this work was published in~\cite{eddeep}. Compared with that conference paper, the present article introduces more principled model choices, more advanced augmentation schemes, and a substantially extended evaluation, including a comprehensive suite of quantitative metrics and an assessment of generalisation on an out-of-domain dataset.

\section{Modelling of eddy-current-induced distortions}
In this section, we describe the physical origin of eddy-current distortions in diffusion MRI, formulate the geometric distortion model and the associated Jacobian intensity modulation, and explain how between-volume head motion is accounted for. Together, these define the constrained transformation that is used for correction.

\subsection{Origin of eddy-current-induced distortions}

The geometric effects of unwanted magnetic fields superimposed on the main field $B_0$ are part of the general theory of EPI distortions~\citep{chang1992}. In diffusion MRI, these effects arise prominently through eddy currents induced by diffusion-encoding gradients~\citep{jezzard1998,jones2010}. Because diffusion-encoding gradients are time-varying magnetic fields, they induce a response in the conducting surfaces of the scanner hardware in the form of swirling electrical currents, hence the name eddy currents. These currents in turn generate residual magnetic fields that interfere with the spatial encoding during image readout. These effects are usually limited in conventional MRI, where short and relatively weak imaging gradients allow partial compensation between the fields induced during the ramp-up and ramp-down phases. In diffusion MRI, however, because the sequence relies on stronger and longer gradient pulses to achieve the required $b$-values, eddy-current-induced distortions can reach an amplitude of several millimeters.

\subsection{Eddy distortion model}
\label{eddydisto}
The eddy-current-induced distortions are typically approximated by a simplified geometric model, from which a correcting transformation can be estimated. Here we describe the model used in this work and how it is applied in practice.
The residual fields caused by eddy currents produce distortions predominantly along the phase-encoding direction (PED), where the bandwidth is low, while their effects along the other directions are negligible~\citep{chang1992}. Since these residual fields are induced by the diffusion-encoding gradients, the distortion varies with the amplitude and orientation of the applied diffusion encoding. Successive diffusion-weighted volumes are therefore affected differently, leading to between-volume misalignment. The effect is specific to diffusion-weighted images, whereas $b=0$ images are unaffected.
As a first order approximation, originally formulated in 2D, slice by slice~\citep{jezzard1998}, the induced displacement along the PED may be expressed as a linear combination of the spatial coordinates. In this formulation, the distorting effects of the eddy currents can be interpreted according to the direction of the residual field: a residual gradient along the read direction produces a shear; a residual gradient along the PED produces a stretch or compression; and a residual gradient along the slice direction shifts each slice uniformly along the PED, like a constant field offset would, but by an amount proportional to the slice position, i.e. a shear between the slice and phase-encoding directions in 3D~\citep{alexander1997,jones2010}.
However, empirical studies have shown that this linear approximation is not always sufficient and that higher-order terms may be required~\citep{rohde2004,andersson2016}. In practice, a quadratic model has proved to be a good compromise between flexibility and parsimony, and is the model adopted by widely used correction tools such as FSL~Eddy~\citep{andersson2016} and Tortoise~\citep{tortoise,tortoisev4}.
Since the residual fields satisfy the Laplace equation within the imaging volume, they can be expanded in spherical harmonics~\citep{barmet2008}, and so can the induced displacement along the PED. Tortoise uses such a harmonic basis, whereas FSL~Eddy uses the full quadratic polynomial basis, which contains one additional, non-harmonic term, $x^2+y^2+z^2$. We follow the latter in this work.

Distortions also have consequences on the intensity of the reconstructed image, the geometric changes are accompanied with intensity modulation. Below, we formulate the distortion model used in this work along with the Jacobian intensity modulation technique used to describe and account for these intensity changes.

\subsubsection{Uni-directional quadratic distortion model}
\label{distomodel}

\begin{figure}[!h]
    \centering
    \includegraphics[width=0.8\linewidth]{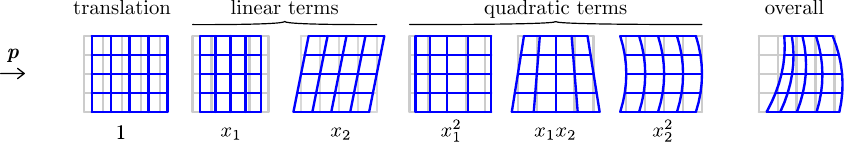}
    \caption{An illustration of the uni-directional quadratic geometric deformation model in 2D.}
    \label{defoquad}
\end{figure}

Let $\boldsymbol{x}\in\mathbb{R}^3$ denote the undistorted spatial coordinates and $\boldsymbol{y}=E(\boldsymbol{x})$ the coordinates distorted by the eddy-current-induced transformation $E$. Let $\boldsymbol{p}\in\mathbb{R}^3$ with $\|\boldsymbol{p}\|=1$ be the vector defining the PED (e.g. $\boldsymbol{p}=(0\ 1\ 0)^\top$ if the PED is along the second axis). The quadratic distortion model can be written as:
\begin{equation}
\boldsymbol{y}=E(\boldsymbol{x}) = \boldsymbol{x}+\left(\boldsymbol{x}^\top \boldsymbol{Q}\boldsymbol{x}+\boldsymbol{L}\boldsymbol{x}+t\right)\boldsymbol{p}
\end{equation}
where $\boldsymbol{Q}\in \mathbb{R}^{3\times 3}_{\mathrm{sym}}$, $\boldsymbol{L}\in \mathbb{R}^{1\times 3}$ and $t\in \mathbb{R}$.
This uni-directional (constrained along the PED) global transformation has 10 degrees of freedom (dof) in 3D (1 for the translation, 3 for the linear terms and 6 for the quadratic terms).
As an illustration, Fig.~\ref{defoquad} shows the 2D version of this distortion model (which has only 6 dof). A harmonic model would impose $\mathrm{tr}(\boldsymbol{Q})=0$, thus removing one dof.

\subsubsection{Jacobian intensity modulation}
\label{jacmodul}

Wherever a transformation $T$ locally compresses space, the signal from a given volume element is concentrated into a smaller reconstructed volume and the intensity increases; conversely, wherever $T$ locally expands space, the same signal is spread over a larger volume and the intensity decreases. Formally, for $I$ denoting the ideal undistorted image and $I'$ the acquired distorted one, following~\cite{chang1992}:
\begin{equation}
    I'\circ T(\boldsymbol{x}) = \frac{1}{|\det \boldsymbol{J}_T(\boldsymbol{x})|} \cdot I(\boldsymbol{x})
\end{equation}
where $\boldsymbol{J}_T(\boldsymbol{x})$ denotes the Jacobian matrix of $T$ at $\boldsymbol{x}$, and $\det$ is the determinant. Values of $|\det \boldsymbol{J}_T(\boldsymbol{x})|$ below 1 correspond to local compression and values above 1 to local stretch. This shows that simple resampling of the distorted image back onto the reference grid is insufficient: the intensity must also be compensated following: $I(\boldsymbol{x}) = |\det \boldsymbol{J}_T(\boldsymbol{x})| \cdot I'\circ T(\boldsymbol{x})$.

\subsection{Accounting for between-volume head motion}
\label{headmotion}
The modelling of eddy-current distortion must take into account head motion. As acquiring a 3-D volume (as a stack of 2-D slices) with EPI takes a few seconds, it is common to consider within-volume head motion as negligible and to only model between-volume head motion. The between-volume head motion can be modelled as a rigid transformation (6 dof in 3D). In the presence of between-volume head motion, the eddy-current distortion must be combined with the rigid transformation of the head to form the complete transformation that the correction aims to undo.

\begin{figure}[!h]
    \centering
    \includegraphics[width=0.8\linewidth]{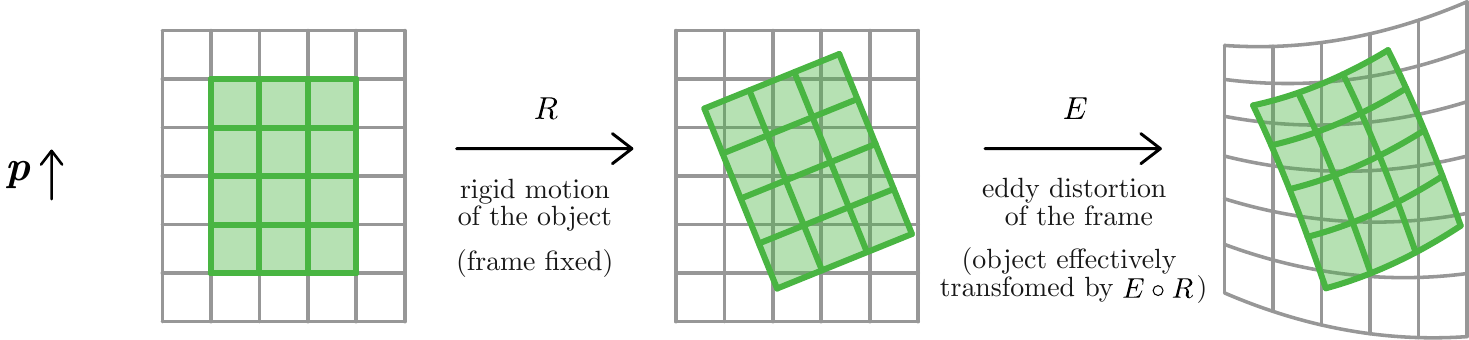}
    \caption{Rigid motion of an object in a fixed reference frame followed by eddy distortion of the whole frame.}
    \label{deforigquad}
\end{figure}

A key consideration is the order in which the rigid motion $R$ and the eddy distortion $E$ are composed. While the subject's head can move freely in physical space, distortions are image reconstruction artifacts that are anchored to the reference frame defined by the imaging gradients (slice-select, phase-encoding and frequency-encoding), as illustrated in Fig.~\ref{deforigquad}. The transformation, $T$, combining both head motion and eddy distortions therefore acts on spatial coordinates $\boldsymbol{x}$ following: 
\begin{equation}
T(\boldsymbol{x})=E\circ R(\boldsymbol{x})
\end{equation}
Eddy and rigid transformations are kept separate for interpretability, but their parameters are not fully independent: in particular, the eddy translation $t$ and the rigid-body translation along the PED are geometrically confounded. Since the rigid component is volume-preserving, the Jacobian determinant used for intensity modulation (Section~\ref{jacmodul}) simplifies to $\det \boldsymbol{J}_T(\boldsymbol{x})=\det \boldsymbol{J}_E(R(\boldsymbol{x}))$.

\section{The Eddeep framework}

\begin{figure}[!h]
    \centering
    \includegraphics[width=\linewidth]{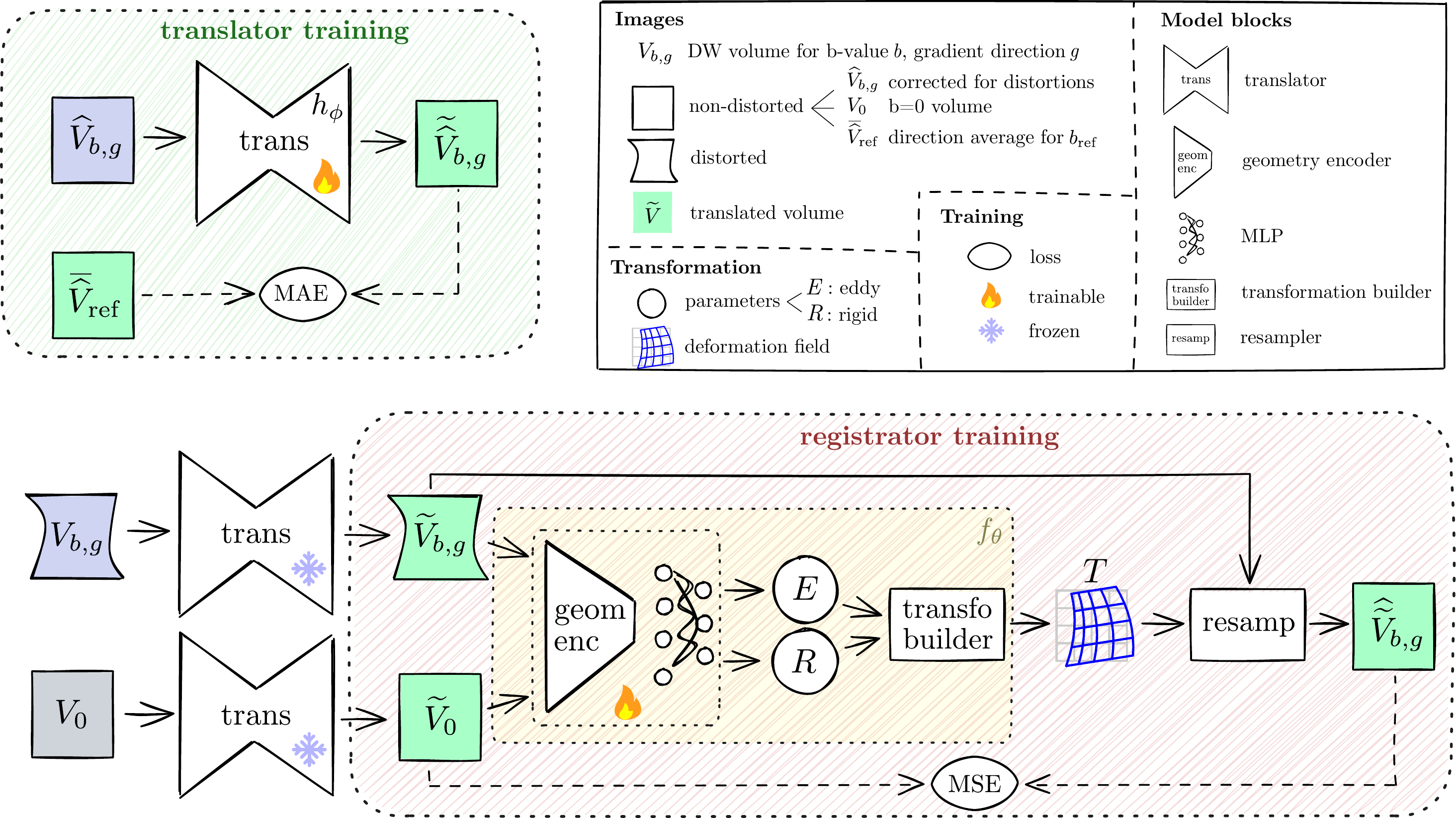}
    \caption{\textbf{Overview of \emph{Eddeep} during training.} The two models are trained sequentially. \textbf{Translator} (top)\textbf{:} each previously-corrected volume ($b=0$ or diffusion-weighted) is passed through a 3D U-Net and mapped to a common target appearance. \textbf{Registration model} (bottom)\textbf{:} with the translator frozen, a translated $b=0$ reference and a translated DW volume are fed to a CNN geometry encoder followed by an MLP that regresses eddy distortion and head-motion parameters. These are assembled into the final transformation, which warps the moving volume.}
    \label{diagmodel}
\end{figure}

We propose a two-stage approach to correct for eddy-current induced distortion and between-volume head motion in which appearance and geometry are decoupled. First, a translator (see Section~\ref{translator}) maps any input diffusion-weighted (DW) or $b=0$ volume to a standardised appearance, restoring correspondences between volumes. Second, a registration model (see Section~\ref{registrator}) estimates the eddy-current distortion and head-motion parameters that align each translated DW volume to a translated, undistorted $b=0$ reference, and applies the correction. An overview is shown in Fig.~\ref{diagmodel}. The coupling between the two models (see Section~\ref{coupling}) and the augmentation strategies used during training (see Section~\ref{aug}) are discussed below.

\subsection{Coupling}
\label{coupling}

Since this approach involves two models, we must decide whether they should be trained jointly. Joint end-to-end optimisation would admit a degenerate solution: in the absence of explicit supervision on the translation output, the translator could collapse to mapping all inputs to zero, yielding a trivially perfect similarity score for the registration objective. 
The translator and registration model are therefore trained sequentially: the translator is trained first, and its weights are then frozen while training the registration model. Freezing the translator provides the registration model with a stable input distribution and, by separating the two training stages, each model has a clear and well-defined optimisation objective.

\subsection{Image translation model}
\label{translator}

The role of the translator is to map any DW or $b=0$ image to a standardised appearance that restores between-volume correspondences while preserving geometry. This is necessary because the registration step that follows must compare volumes acquired at different $b$-values and gradient directions, whose native intensity patterns are otherwise too dissimilar for reliable direct alignment. The architecture and training workflow of the translator are illustrated Fig~\ref{diagmodel}-top. Here we detail the design of the translator, including the choice of target contrast, training methodology and inference usage. 

\subsubsection{Training objective}
Training uses paired previously-corrected data so that the learning problem is purely one of contrast mapping, without geometric mismatch between input and target. Let $h_{\phi}$ denote the translator model with weights $\phi$. For an input previously-corrected volume $\hat V_{b,\boldsymbol{g}}$, of $b$-value $b$ and gradient direction $\boldsymbol{g}$, and the corresponding subject-specific target $\bar V_{\mathrm{ref}}$, the translator is trained by minimising
\begin{equation}
    \phi^{*} = \operatorname*{arg\,min}_{\phi}\ \mathbb{E}_{(\hat V_{b,\boldsymbol{g}},\,\bar V_{\mathrm{ref}})}\;
    \mathcal{R}\bigl(h_{\phi}(\hat V_{b,\boldsymbol{g}}),\,  \bar V_{\mathrm{ref}}\bigr).
\end{equation}

For $\mathcal{R}$, the reconstruction loss, we use MAE ($L_1$) rather than MSE ($L_2$) as in~\cite{eddeep}. The two differ in how they handle uncertainty~\citep{bishop2006}: MSE minimisation pushes the network toward the mean of all plausible outputs, potentially producing blurry predictions, whereas MAE minimisation pushes it toward the median, which is more robust and can better preserve sharp edges. The same translator is applied to both DW and $b=0$ volumes, mapping them to the same target appearance.

\subsubsection{Architecture}

Since the translation must not introduce any spatial shift, a constrained paired supervised approach is preferred to unpaired alternatives such as cycleGAN~\citep{zhu2017}. In the conference version~\citep{eddeep}, we used a conditional GAN in a pix2pix~\citep{isola2017} style. For this task, however, we do not need to generate high-frequency details or photorealism: the precise appearance of the translated image matters less than its geometric fidelity and the absence of hallucination. We therefore simplify the translator to a plain 3D U-Net~\citep{unet} trained with the regression objective above, which also yields more stable training.

\subsubsection{Training and inference usage}

\textbf{Training:} The translator requires previously-corrected data on both sides of the mapping. As with target construction, the inputs must be previously-corrected so that they share the same geometry as the target, keeping training as a pure contrast-learning problem with no geometric mismatch to confound the loss.

\textbf{Inference:} The translator is instead applied to distorted volumes, whose translated outputs serve as inputs to the registration model. This mismatch between training and inference is tolerable because eddy distortions, while significant enough to affect downstream modelling and analysis, are small in amplitude relative to the variations the U-Net generalises over; spatial augmentation applied during training further closes this gap.

\subsubsection{Translation target}
The choice of the translation target is guided by the following observations:
\begin{itemize}
   \item For a given $b$-value, averaging volumes over gradient directions (assuming approximately uniform sphere sampling) removes orientation-specific signal attenuation, yielding a target with no direction dependency.
   \item It is easier to remove structure that is present in the input than to generate structure that is absent: translating from a low $b$-value (where outer CSF produces a bright boundary) to a higher one (where CSF is attenuated) requires suppression rather than hallucination.
   \item At moderately high $b$-values (typically 700--3000\,s/mm²), the CSF signal is fully attenuated to noise level while tissue SNR remains sufficient to preserve the edges that drive alignment.
\end{itemize}
Our translation target is therefore the image obtained by averaging all diffusion volumes acquired at a moderate $b$-value $b_\mathrm{ref}$ over all gradient directions. This target is practically built per subject by selecting a $b_\mathrm{ref}$ in the 700--3000\,s/mm² range from the acquisition protocol, correcting the associated volumes with an external tool (e.g.\ FSL~Eddy), averaging them, and applying bias field correction.

The translation target is used only during training for supervision. At inference, the translator operates directly on individual diffusion-weighted or $b=0$ volumes, without requiring averaging, pre-correction, or uniform gradient sampling. The assumptions underlying target construction apply solely to training and do not constrain model deployment.

\subsection{Image registration model}
\label{registrator}

The role of the registration model is to estimate the geometric transformation that aligns a translated distorted DW volume to a translated undistorted $b=0$ reference. It jointly accounts for eddy-current distortion and between-volume head motion. The architecture and training workflow of the registration model are illustrated Fig~\ref{diagmodel}-bottom. Here we detail the design of the registration model, training methodology and inference usage. 

\subsubsection{Training objective}

Let $f_{\theta}$ denote the registration model with parameters $\theta$. Following recent learning-based registration approaches~\citep{voxelmorph1,devos2017}, we train $f_\theta$ to take a pair of images as input and predict the transformation that aligns them in a single forward pass. The training is unsupervised, so no ground-truth transformations are required. 

The model weights are optimised by maximising an intensity-based similarity criterion $\mathcal{S}$ between the reference image and the moving image warped by the predicted transformation.

In our setting, the input pair consists of a translated $b=0$ reference $\widetilde{V}_0$ and a translated DW volume $\widetilde{V}_{b,\boldsymbol{g}}$ of $b$-value $b$ and gradient direction $\boldsymbol{g}$, from the same acquisition. The registration model predicts a transformation $T = f_\theta(\widetilde{V}_0, \widetilde{V}_{b,\boldsymbol{g}})$ that follows the transformation model introduced in Section~\ref{distomodel}: $T = E \circ R$ , where $E$ encodes the unidirectional quadratic eddy distortion (10 dof) and $R$ accounts for between-volume head motion (6 dof). The training objective is therefore:
\begin{equation}
    \label{obj_corr}
    \theta^* = \operatorname*{arg\,min}_{\theta}\; 
    \mathbb{E}_{(\widetilde{V}_0,\,\widetilde{V}_{b,\boldsymbol{g}})}\; 
    \mathcal{S}\!\left(\widetilde{V}_0,\; 
    \widetilde{V}_{b,\boldsymbol{g}} \circ f_\theta(\widetilde{V}_0,\,\widetilde{V}_{b,\boldsymbol{g}})\right)\ ,
\end{equation}
Because the translator equalises contrast across volumes, a very simple similarity measure such as MSE can be used, regardless of the $b$-value and gradient direction of the moving volume.

\begin{remark}
As defined in Section~\ref{distomodel}, $\boldsymbol{y} = T(\boldsymbol{x})$ is the forward geometric mapping from a coordinate $\boldsymbol{x}$ in the undistorted image to its corresponding location $\boldsymbol{y}$ in the distorted image. Therefore, reconstructing the corrected image $\widehat{V}$ on the undistorted grid is achieved by backward resampling of the distorted image $V$, following: $\widehat{V}(\boldsymbol{x}) = V(T(\boldsymbol{x}))$. 
Hence, the quantity to be estimated is the forward transformation $T$, not its inverse.
\end{remark}

\subsubsection{Architecture}
\label{architecture}

Since the transformation model is low-dimensional and global (see Section~\ref{distomodel}), a parameter-regression approach is more appropriate than a dense deformation decoder. We therefore follow a strategy in the same vein as what is used in~\cite{devos2017,devos2019} for affine registration, replacing the decoder part of the U-Net by fully connected layers (a multilayer perceptron, or MLP) that directly regress transformation parameters.

The model is composed of the following blocks:

\begin{enumerate}
    \item \textbf{Input:} A distorted translated DW volume $\widetilde{V}_{b,\boldsymbol{g}}$ (moving) and a translated $b=0$ image $\widetilde{V}_0$ (reference) of a same acquisition are concatenated along the channel dimension.
    
    \item \textbf{Geometry encoder:} A CNN encoder extracts features at progressively coarser spatial resolutions, capturing local anatomical detail as well as global structural context. No decoder or skip connections are needed, since the task is low-dimensional parameter regression.

    \item \textbf{Dense layers:} The flattened encoder output is passed through the MLP, which aggregates information over the entire volume before regressing the global transformation parameters. The final layer consists of five parallel linear output heads, equivalent to a single linear layer with 16 outputs, grouping the parameters by type: eddy translation ($e_1$), eddy linear ($e_2,e_3,e_4$), eddy quadratic ($e_5,\dots,e_{10}$), rigid translation ($r_1,r_2,r_3$), and rigid rotation ($r_4,r_5,r_6$).

    The along-PED translation of the rigid transformation and the eddy translation $e_1$ are confounded (Section~\ref{headmotion}), but separating them here eases recovery of either transformation alone if needed, without affecting the estimation of $T$.

    \item \textbf{Transformation builder:} The parameters output by the MLP are assembled into the eddy transformation $E$ and rigid transformation $R$, following the transformation model defined in Section~\ref{distomodel}. Briefly, the matrix exponential of a skew-symmetric matrix is used to guarantee a proper rotation matrix for any unconstrained parameter values, and the eddy parameters directly build the unidirectional quadratic transformation (see Appendix~\ref{appendix} for full details). 
    Both transformations are expressed relative to the image centre, so that zero parameters yield the identity and small parameter values yield small transformations centred on the brain, rather than, e.g., large displacements far from a corner origin. Following Section~\ref{headmotion}, the composition is ordered as $T = E \circ R$.
    
    \item \textbf{Resampler:} The overall transformation is used to warp the moving image via backward resampling, in a way that differs between training and inference (see Section~\ref{traininf_corr}).
\end{enumerate}

\subsubsection{Training and inference usage}
\label{traininf_corr}

\textbf{Training:} The predicted transformation is applied to the translated DW volume: $\widehat{\widetilde{V}}_{b,\boldsymbol{g}}(\boldsymbol{x})=\widetilde{V}_{b,\boldsymbol{g}} \circ T(\boldsymbol{x})$, which is then compared to the translated reference $\widetilde V_0$ using the similarity metric $\mathcal{S}$, following Eq~\ref{obj_corr}.

\textbf{Inference:} The predicted transformation is instead applied to the original DW volume and, following Section~\ref{jacmodul}, Jacobian intensity modulation is used to compensate for signal pile-up and stretch:
$\widehat{V}_{b,\boldsymbol{g}}(\boldsymbol{x})=|\det \boldsymbol{J}_T(\boldsymbol{x})|\cdot V_{b,\boldsymbol{g}}\circ T(\boldsymbol{x})$ where $\det \boldsymbol{J}_T(\boldsymbol{x})$ is the determinant of the Jacobian of the transformation $T$ at $\boldsymbol{x}$. This yields a corrected DW volume mapped to the undistorted geometry referenced to $b=0$.

\subsection{Augmentation strategies}
\label{aug}

The purpose of augmentation is to broaden the training distribution to improve the generalisability of both models. Contrast augmentations are designed to cover a wider range of acquisition protocols and imaging conditions, while spatial augmentations expose the model to more head orientations and anatomical variability. Although the proposed augmentations are inspired by physical principles, exact realism is not required. The augmentation schemes compatible with the translation and registration models differ, and are detailed below.

\subsubsection{Translator}
\label{aug_trans}

\begin{itemize}
\item \textbf{Contrast:}

\begin{itemize}
\item \textbf{\textit{b}-value synthesis:} Since the data is previously corrected, one can synthesise signal at new effective $b$-values from a corrected diffusion-weighted volume $\widehat{V}_{b,\boldsymbol{g}}$ and a $b=0$ volume $\widehat{V}_0$ of the same subject via interpolation or extrapolation under a mono-exponential signal model:
\begin{equation}
\widehat{V}_{\alpha b,\boldsymbol{g}} \approx \exp\big(\alpha \log(\widehat{V}_{b,\boldsymbol{g}}) + (1-\alpha)\log(\widehat{V}_0)\big), \quad \alpha \in [0,+\infty).
\end{equation}
For $0 \leq \alpha \leq 1$ this interpolates between $b=0$ and the acquired contrast; for $\alpha > 1$ it extrapolates to a higher effective diffusion weighting. Although based on a simplistic signal model, this approximation is sufficient for contrast augmentation, at least for moderately high $b$-values, and is inexpensive to compute on the fly in the dataloader.

\item \textbf{Bias field:} A synthetic random bias field $B$ is generated as a smooth low-frequency scalar field (typically through low-resolution random Gaussian noise, upsampled to the image resolution and smoothed with a Gaussian kernel) and applied multiplicatively: $\widehat{V}_\mathrm{bias}=(1+B)\,\widehat{V}$. The result of applying another smooth multiplicative field on top of a pre-existing one remains smooth, effectively simulating a larger or differently shaped inhomogeneity.

\item \textbf{Noise:} Rician noise $N$ is added to the translator input to improve robustness to varying SNR: $\widehat{V}_\mathrm{noise}=\widehat{V}+N$. Strictly speaking, adding Rician noise to an already noisy image does not reproduce a lower-SNR Rician distribution, and multi-coil acquisitions lead to more complex noise statistics further disrupted by previous correction. In practice however, this approximation is adequate to introduce useful noise diversity.

\end{itemize}

These are combined in the following way: $\widehat{V}_\mathrm{aug} = (1+B)\,\widehat{V}_{\alpha b,\boldsymbol{g}}+N$. They are applied to the translator input only, while the translation target $\bar V_{\mathrm{ref}}$ is kept unchanged.

\item \textbf{Spatial:} The same composition of a random global affine transformation and a random diffeomorphic deformation field is applied to both the input and the translation target. This preserves voxel-wise correspondence and keeps the learning problem purely one of contrast mapping.

\end{itemize}

\subsubsection{Registration model}
\label{aug_corr}
\begin{itemize}
\item \textbf{Contrast:} No contrast augmentation is used for the registration model. Its inputs are the translated volumes $\widetilde{V}_0$ and $\widetilde{V}_{b,\boldsymbol{g}}$, and the role of the translator is precisely to reduce contrast variability across $b$-values and gradient directions. 

\item \textbf{Spatial:} Any spatial augmentation must preserve compatibility with the transformation relating the moving and reference volumes. This strongly restricts the class of admissible augmentations, and the set of safe transformations may be narrower than implied in the conference version~\cite{eddeep}. In particular, transformations known to be safe include arbitrary translations and rotations about the PED, which may be applied independently to the reference and moving volumes, as well as a common isotropic scaling applied to both images. More general rotations, anisotropic scaling, shear, and deformable transformations would in general produce relative transforms outside the assumed model.

\end{itemize}

\section{Evaluation}

In this section, we evaluate \emph{Eddeep} in terms of both translation quality and correction quality. We first lay out the experimental design: the datasets used for in-domain and out-of-domain evaluation (Section~\ref{dataset}), the implementation details and evaluated variants of our models (Section~\ref{eddepmodels}), the FSL~Eddy baseline (Section~\ref{baseline}), and the masks and notation shared across all metrics (Section~\ref{masks}). We then assess the translation stage on its own (Section~\ref{eval_trans}), before evaluating the full correction framework through a range of complementary quantitative metrics (Section~\ref{eval_corr}). Both stages are assessed on in-domain and out-of-domain data, providing an assessment of accuracy, robustness and generalisation. Finally, we compare the processing times of \emph{Eddeep} and FSL~Eddy (Section~\ref{proctime}).

\subsection{Experimental design}

\subsubsection{Datasets}
\label{dataset}
We evaluated the proposed method on two diffusion MRI datasets: UK Biobank, used for in-domain training, validation, and testing, and Memodyn, used exclusively for out-of-domain evaluation.
\begin{itemize}
\item \textbf{UK Biobank (in-domain):} The UK Biobank (UKB) dataset~\citep{miller2016,alfaro2018} is a large-scale population cohort acquired on a 3T Siemens Skyra scanner. The multi-shell dMRI protocol consists of 5 $b=0$ volumes and 50 gradient directions at each of $b=1000$ and $b=2000$\,s/mm², voxel size is $2\times2\times2$\,mm, matrix $104\times104$, 72 slices, TE\,=\,92\,ms, TR\,=\,3600\,ms.

\par\smallskip
Both the translator and the registration model were trained exclusively on UKB data, using 80 subjects each. Their training sets partially overlapped, with 40 subjects shared between them. The overlap allows the total training data to be reduced, but is kept partial on purpose: exposing the corrector to subjects unseen by the translator during training ensures it learns to handle imperfect translations, which is representative of inference conditions. 

Validation used 14 subjects per model (7 shared). The 31 held-out UKB subjects used for in-domain testing have no overlap with either training or validation split.

\item \textbf{Memodyn (out-of-domain):} The Memodyn dataset~\citep{stee2023} (not yet publicly available) was acquired on a 3T Siemens Magnetom Prisma scanner as part of a study investigating memory learning and consolidation. The multi-shell dMRI protocol consists of 13 $b=0$ volumes and diffusion-weighted images at $b=650$, $1000$ and $2000$\,s/mm² with 15, 30 and 60 gradient directions respectively, voxel size is $2\times2\times2$\,mm, matrix $96\times96$, 70 slices, TE\,=\,70.2\,ms, TR\,=\,4070\,ms.

\par\smallskip
Its 50 subjects were entirely unseen during training, constituting the out-of-domain evaluation set.
\end{itemize}

The diffusion encoding parameters and data splits of the two datasets are summarised in Table~\ref{tab_dataset}\footnote{The shell illustrations were obtained using: \url{http://www.emmanuelcaruyer.com/q-space-sampling.php}.}. In both cases, eddy-current correction was performed on raw images that had not been previously corrected for susceptibility-induced distortions. For both, the PED is along the antero-posterior direction (AP). The two datasets differ in their level of eddy distortions: UKB data acquired on the older Skyra platform exhibit visually more pronounced distortions than the Prisma-acquired Memodyn data, consistent with differences in gradient performance and eddy-current compensation hardware. This is also reflected in the translated jitter metric computed on uncorrected images, which is markedly higher for UKB (see Fig.~\ref{trans_jitter_map}).

\begin{table}
\centering
\includegraphics[width=\linewidth]{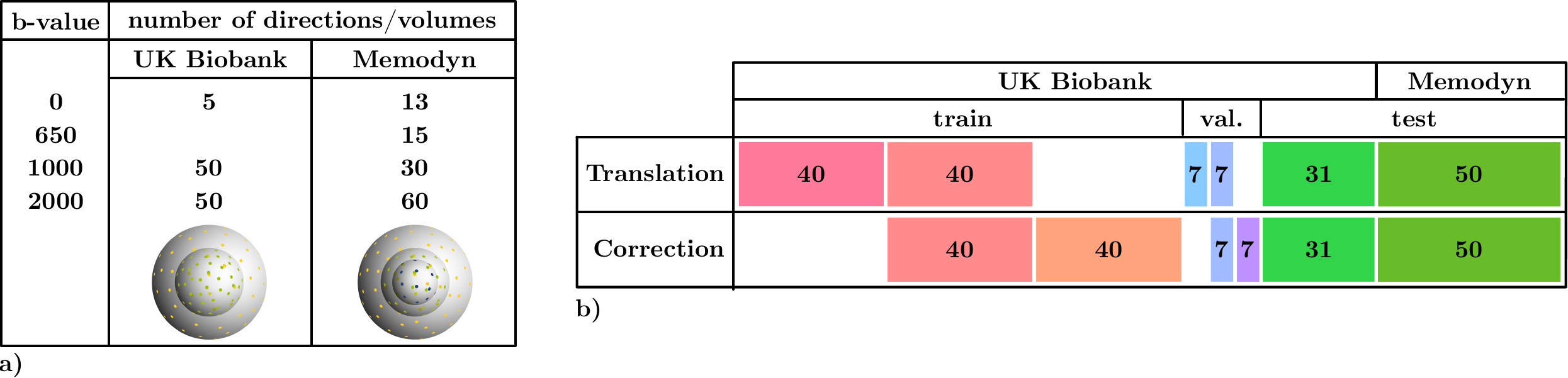}
\caption{Dataset protocols and splits: a) shell $b$-values and number of gradient directions/volumes in each dataset; b) dataset split for training and testing the translation and registration models, with partial overlap in the training and validation splits between both models.}
\label{tab_dataset}
\end{table}

\subsubsection{Eddeep models}
\label{eddepmodels}

\textbf{Implementation details:}
\begin{itemize}

\item \textbf{Translator:}
The translator follows a 3D U-Net~\citep{unet} architecture. The encoder comprises 4 levels with 16, 32, 64 and 128 channels, and the decoder mirrors this structure with 128, 64, 32 and 16 channels. Each level has 2 convolutional blocks with Leaky ReLU activations and $3\times 3 \times 3$ kernel size. Downsampling is performed by max-pooling (factor 2), upsampling by nearest-neighbour interpolation (factor 2), and skip connections by concatenation. The final layer is a single-channel convolution with no activation. 

During training, the model is applied independently to both the $b=0$ and the DW volume, each mapped to the mean DW image at the target $b$-value, and the two MAE losses are averaged.
Training uses the Adam~\citep{diederik2014} optimiser with learning rate $10^{-4}$ and batch size 4. The selected model is the best according to the validation loss over 300 epochs.

For some of the model variants we evaluate, a discriminator is added; these variants will be indicated by the inclusion of "GAN" in their names. The discriminator is a PatchGAN~\citep{isola2017} CNN encoder with features (16, 32, 64), MSE loss with weight $10^{-2}$ compared to the generator. Together with the generator, it forms a pix2pix~\citep{isola2017}-style conditional GAN, similarly to what is used in~\cite{eddeep}.

To construct the training inputs and targets, the data are previously corrected using FSL~Eddy~\citep{andersson2016}. The reference shell is chosen as $b_{\mathrm{ref}}=2000$. Because the $b$-vectors are uniformly distributed over the sphere in the UKB dataset, averaging the volumes associated with $b_{\mathrm{ref}}$ cancels direction-dependent attenuation patterns. The resulting mean image is then bias-field corrected using ANTs N4~\citep{tustison2010}. Input volumes are normalised to $[0,1]$ by clipping at the quantile 0.999 of the intensity distribution and dividing by that quantile. 

\item \textbf{Registration model:}
The registration model takes as input a pair of translated volumes, namely a translated $b=0$ reference and a translated DW volume generated by the frozen translator. Its geometry encoder is a 3D CNN with 5 levels, containing 16, 28, 56, 75 and 128 channels. Each level has 2 convolutional blocks with Leaky ReLU activations and $3\times 3 \times 3$ kernel size. Downsampling is performed by max-pooling (factor 2). The flattened encoder output is passed through to a 2-layer MLP with 128 and 64 nodes, using Leaky ReLU activations. The MLP output is split into 5 parallel branches with no activation: eddy-current quadratic (6 nodes), linear (3 nodes) and translation (1 node) parameters; rigid rotation (3 nodes) and translation (3 nodes) parameters.
The resulting transforms are composed and applied to the DW volume via trilinear interpolation, using Voxelmorph's~\cite{voxelmorph1} code components, adapted to handle the quadratic distortion model of Section~\ref{distomodel}. 

The training loss is MSE between the corrected translated DW and the translated $b=0$. Training uses the Adam optimiser with learning rate $10^{-4}$ and batch size 4. The selected model is the best according to the validation loss over 200 epochs. 
\end{itemize}

On an NVIDIA RTX~3090 GPU, training took about 22\,h for the translator with 300 epochs and about 13.5\,h for the registration model with 200 epochs. The duration could be shortened for the registration model though, as its loss drops sharply after an initial plateau (Fig.~\ref{seeds}) and then roughly stagnates, so that training could be stopped shortly after this drop is detected. The translator loss on the other hand shows a typical exponential-like decrease. 

The code is available at:~\url{https://github.com/CIG-UCL/eddeep}. Training-run variability of the registration model is assessed in Section~\ref{eval_seeds}.

\textbf{Model variants:}

\begin{itemize}
    \item \textbf{Translation: } We evaluated three variants differing only in the amount of augmentation (both in contrast and spatial) applied during training: \texttt{trans}, with no augmentation; \texttt{trans+}, with augmentation applied with probability $0.5$; and \texttt{trans++}, with augmentation applied with probability $1$. A GAN version of \texttt{trans+} was also evaluated.

    \item \textbf{Full Eddeep (translation+correction): } the \emph{Eddeep} variants combine translation and registration in sequence. The correction model of \texttt{eddeep} was trained using the frozen translator \texttt{trans}, whereas the correction model of \texttt{eddeep+} was trained using the frozen translator \texttt{trans+}. No augmentation was used during training of the registration model.

    All \emph{Eddeep} variants use a quadratic unidirectional transformation model, Jacobian intensity modulation, and trilinear interpolation.
\end{itemize}

\subsubsection{Baseline method: FSL Eddy}
\label{baseline}
As a baseline, we used FSL Eddy~\citep{andersson2016} (\texttt{eddy\_openmp}) from the FSL suite~\citep{jenkinson2012} (version 6.0.5.2), which is a standard state-of-the-art tool for eddy-current distortion and motion correction in diffusion MRI. For comparability, we used the same correction setting as in \emph{Eddeep}: a quadratic unidirectional transformation model with Jacobian intensity modulation and trilinear interpolation. Outlier replacement and slice-to-volume motion correction are disabled. 
For the processing time evaluation in Section \ref{proctime} only, \texttt{eddy\_cuda10.2} GPU version was also used.

\subsubsection{Common masks and notation}
\label{masks}
For the evaluation, we computed a set of common masks to be used for quantification of performance and visualisation. Here, we define these masks and additional notation to be used throughout the experiments.

To compute model performance in relevant regions of the image, we utilise a set of brain masks. $\mathcal{X}_\mathrm{brain}$ denotes the set of voxel locations in the brain mask, obtained with SynthStrip~\citep{hoopes2022} on the translation target images (i.e., the $b=2000$ images averaged over all gradient directions). $\mathcal{X}_\mathrm{rim}$ denotes the set of voxel locations in a rim at the outer edge of the brain, obtained as the difference between a version of the brain mask dilated by 2 iterations and a version eroded by 4 iterations. In addition, to highlight that the image translation does not alter the geometry of the brain, in Fig.~\ref{trans_fig}, we computed a tighter brain mask using SynthSeg~\citep{billot2023} by merging all non-CSF labels. 

We denote by $I$ the full 4D acquisition comprising multiple $b=0$ and diffusion-weighted 3D volumes, by $I_v$ the volume indexed by $v$, and by $\boldsymbol{x}$ the voxel location. The translation target is denoted as $I_\mathrm{ref}$. For a given shell $b$, $\mathcal{V}_b$ denotes the corresponding set of volume indices. Finally, $s_0$ denotes a scaling factor defined as the average signal within $\mathcal{X}_\mathrm{brain}$ in the first uncorrected $b=0$ volume; this is used to normalise signal-based metrics and improve comparability across subjects and datasets.

\subsection{Quality of the translation}
\label{eval_trans}

We first inspect the outputs visually (Section~\ref{eval_visual}), then quantify them with two complementary measures: voxel-wise intensity agreement (Section~\ref{eval_ae}) and preservation of local structure (Section~\ref{eval_ssim}). The four translator variants are compared on both datasets.

\subsubsection{Visual assessment}
\label{eval_visual}

\begin{figure}[!h]
    \centering
    \includegraphics[width=0.9\linewidth]{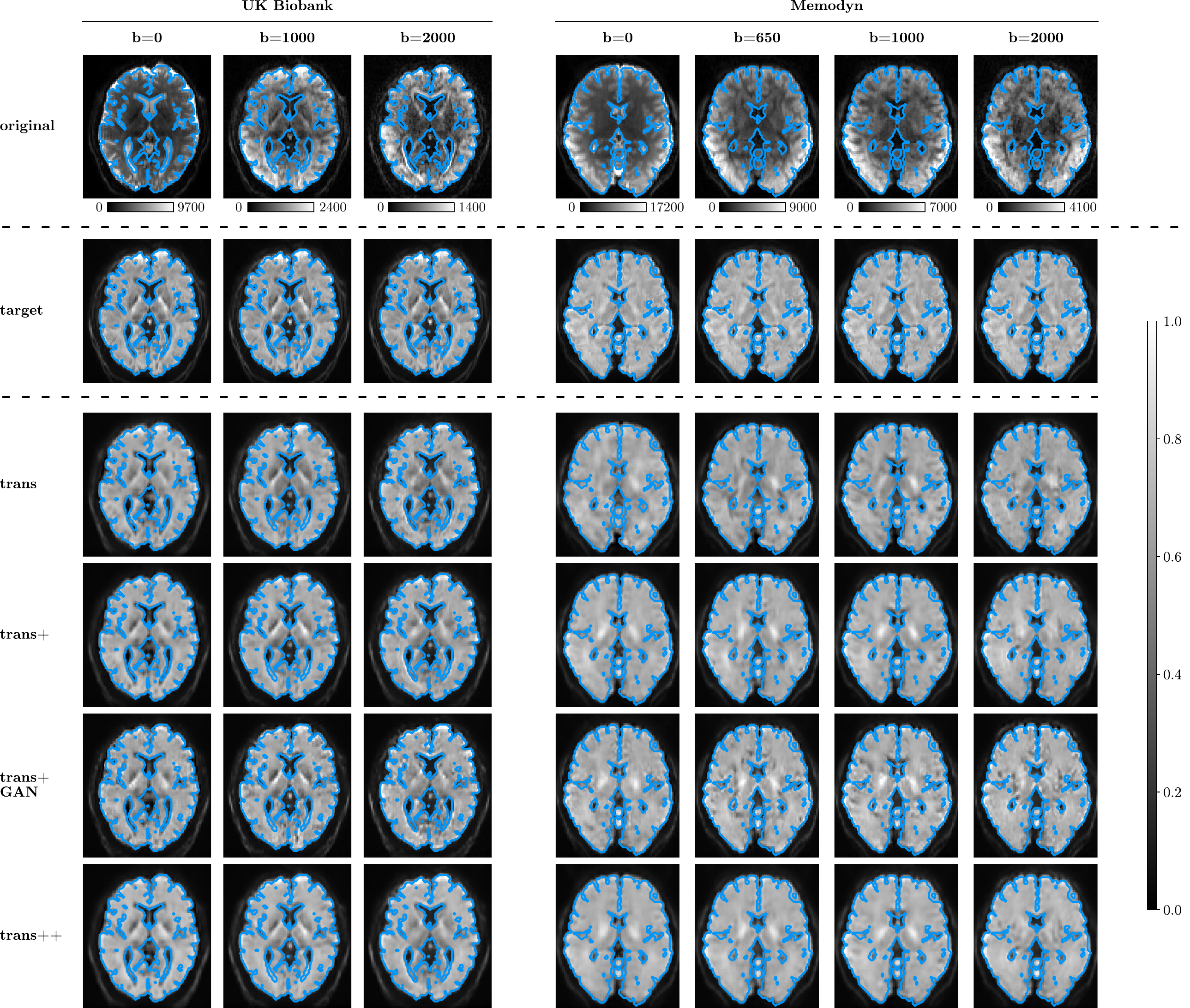}
    \caption{\textbf{Translation outputs.} Representative examples on both datasets for input volumes acquired with different $b$-values and $b$-vectors. The original images are shown together with the translation target and the outputs of the different translator variants. A tight brain mask is overlaid to assess whether the translation preserved the geometry. The translated volumes are visually close to the target contrast, and the geometry appears preserved.}
    \label{trans_fig}
\end{figure}

Figure~\ref{trans_fig} shows representative translation results for volumes acquired with different $b$-values and $b$-vectors, on both UK Biobank and Memodyn data, together with the target appearance and the outputs of the different translator variants.

All models successfully map the inputs towards a common appearance: they eliminate the outer CSF signal and largely remove the $b$-vector-dependent attenuation patterns, making volumes from different parts of the acquisition visually much more comparable. At the same time, the geometry is well preserved, as highlighted by the superimposed tight brain mask. There are some differences between models in the image quality: the GAN variant tends to reproduce noise patterns, which is not useful for our registration task, \texttt{trans} appears somewhat blurrier on the out-of-domain examples, \texttt{trans++} gives more consistent behaviour between in-domain and out-of-domain data, although at the price of a slight overall blur. By eye, \texttt{trans+} appears to offer the best trade-off between contrast standardisation, sharpness, and robustness. Overall, however, the differences between variants remain relatively modest visually.

\subsubsection{Absolute error}
\label{eval_ae}

We computed the voxel-wise absolute error (AE) with respect to the target and summarised it by the mean absolute error (MAE) within the brain mask. This metric provides a direct measure of translation fidelity in the supervised setting: lower values indicate that the translated output is, on average, closer to the target image intensities. Precisely, AE and MAE are defined as:

\begin{equation}
\mathrm{AE}(I,b) = \underset{v\in \mathcal{V}_b}{\operatorname{mean}}\left|I_v - I_\mathrm{ref}\right|\ ,\quad \mathrm{MAE}(I,b) = \underset{\boldsymbol{x}\in\mathcal{X}_\mathrm{brain}}{\operatorname{mean}}\left(\mathrm{AE}(I,b)(\boldsymbol{x})\right)
\end{equation}

\begin{figure}[!h]
    \centering
    \begin{tabular}{c}
        a) \includegraphics[width=0.8\linewidth]{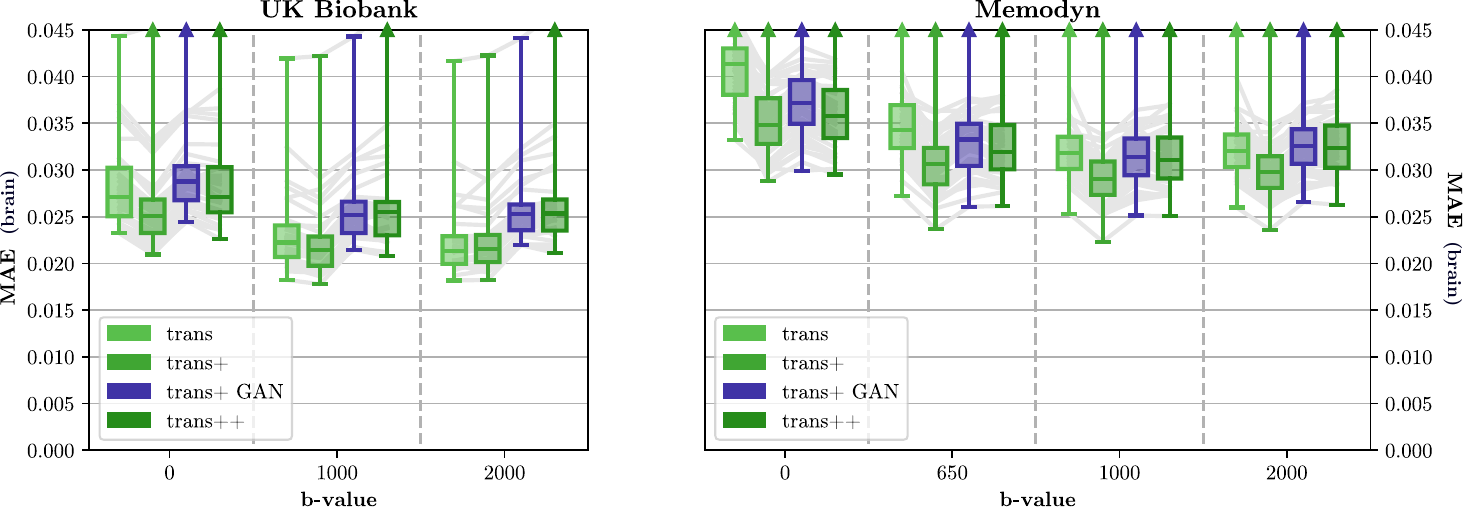}\vspace{0.5cm}\\
        b) \includegraphics[width=0.9\linewidth]{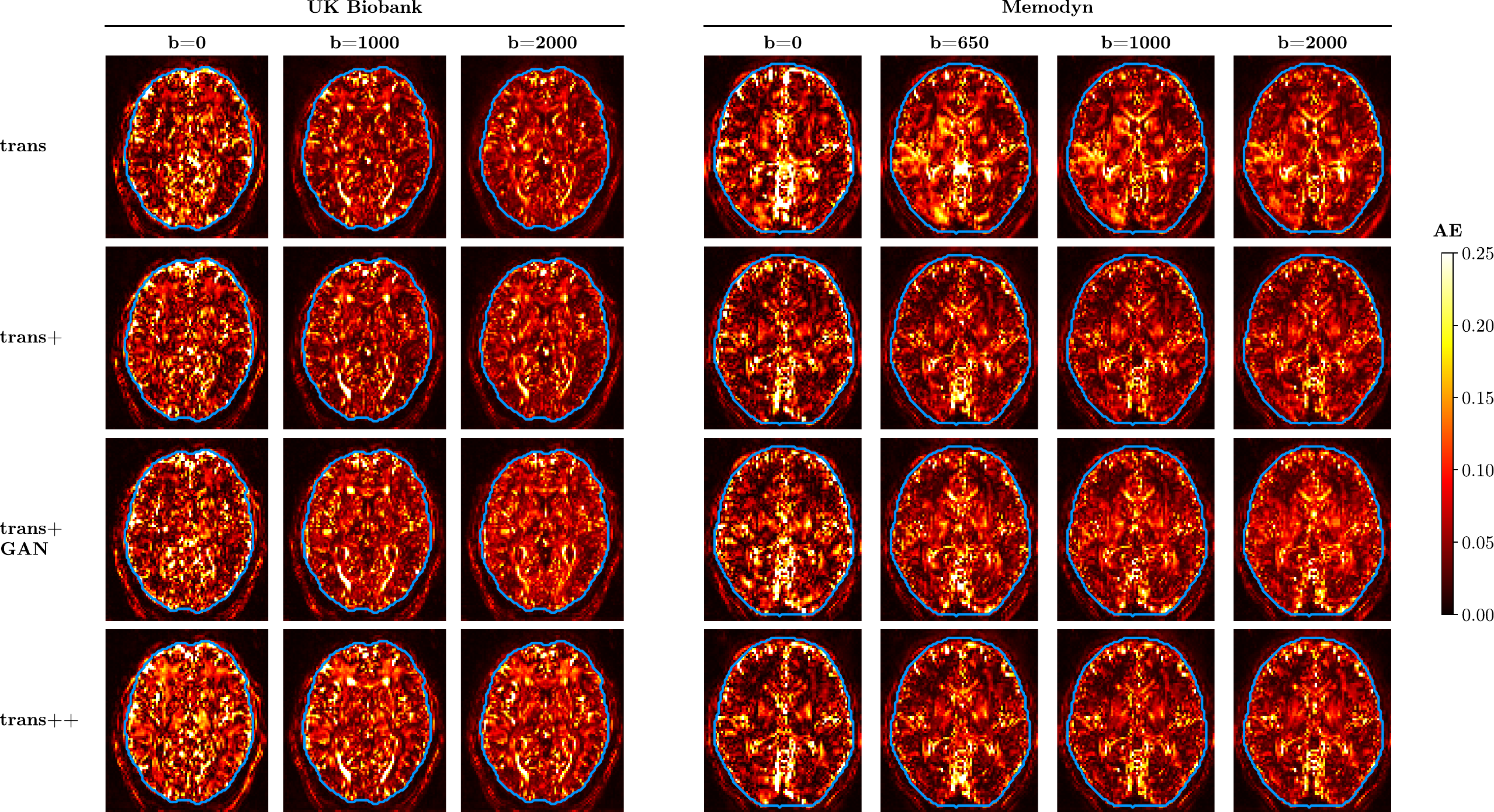}
    \end{tabular}
    \caption{\textbf{Absolute error.} a)~Boxplots of $\mathrm{MAE}(I,b)$ per $b$-value for translated data with respect to the target, with various translator variants; lower is better. b)~Voxel-wise map $\mathrm{AE}$ on a representative subject: the brain mask is outlined in blue. \texttt{trans+} is the best performing model both on in- and out-of-domain datasets.}
    \label{mae_fig}
\end{figure}

\textbf{Results.} Figure~\ref{mae_fig} shows that all variants achieve low error overall. \texttt{trans+} performs best both in-domain and out-of-domain. The other variants, including \texttt{trans+} GAN, give broadly similar results, although trans generalises somewhat more poorly on Memodyn. This is consistent with the visual assessment and supports the benefit of moderate augmentation for robustness.

\subsubsection{Structural similarity}
\label{eval_ssim}

Structural similarity (SSIM) complements absolute error by assessing the preservation of local image structure rather than only voxel-wise intensity agreement. This is particularly relevant here because the role of the translator is not merely to reproduce the target intensities, but to generate images with the correct anatomical organisation for downstream registration. Higher values indicate that the translated images remain structurally closer to the target.

\begin{equation}
\mathrm{SSIM}(I,b) = \underset{v\in \mathcal{V}_b}{\operatorname{mean}}\left(\mathrm{SSIM}(I_v, I_\mathrm{ref})\right)
\end{equation}

\begin{figure}[!h]
    \centering
    \includegraphics[width=0.85\linewidth]{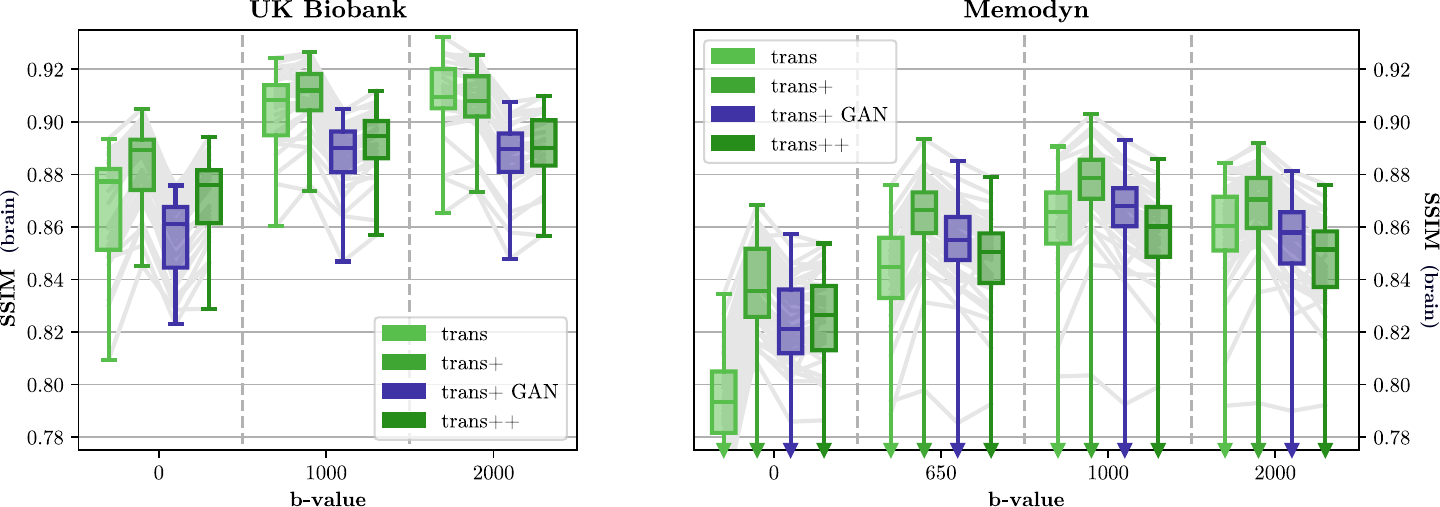}
    \caption{\textbf{Structural similarity} per $b$-value for translated data with respect to the target, with various translator variants; higher is better. \texttt{trans+} is the best performing model both on in- and out-of-domain datasets.}
    \label{ssim}
\end{figure}

\textbf{Results.} Figure~\ref{ssim} shows that \texttt{trans+} again gives the best overall performance, both in-domain and out-of-domain. The other variants, including \texttt{trans+ GAN}, perform similarly overall, whereas \texttt{trans} generalises slightly less well on Memodyn. These results are consistent with the absolute error analysis and further support the use of moderate augmentation.

\subsection{Quality of the correction}
\label{eval_corr}

No ground-truth eddy distortion is available in real data, so no single metric is sufficient. We therefore report several complementary measures, reflecting different potential symptoms of residual misalignment: its direct geometric signature, through between-volume jitter at the brain rim (Section~\ref{eval_jit}) and over the whole brain after translation (Section~\ref{eval_transjit}); its propagation into microstructural modelling, through the DKI fractional anisotropy halo and model residuals (Section~\ref{eval_dki}); the disruption of q-space regularity (Section~\ref{eval_rough}); and its statistical trace, through pairwise mutual information (Section~\ref{eval_mi}). Throughout, the \emph{Eddeep} variants are compared against uncorrected data and the FSL~Eddy baseline, on both datasets. The reported metrics that depend on the $b$-vectors (the DKI ones in Section~\ref{eval_dki} and the signal regularity in Section~\ref{eval_rough}) were computed with the acquired, unrotated $b$-vectors, identically for every method. For each metric, a paired statistical comparison of the \emph{Eddeep} variants with FSL~Eddy, with confidence intervals and effect sizes, is provided in the Supplementary Material.

\subsubsection{Jitter across corrected volumes}
\label{eval_jit}

For a given shell, the voxel-wise intensity standard deviation across volumes quantifies how signal intensity fluctuates when scrolling through the different gradient directions. Regions with strong anisotropic diffusion, such as white matter tracts, are expected to exhibit high variability even when volumes are perfectly aligned, as the signal should vary with gradient direction. However, isotropic regions should show minimal variation. At the outer edge of the brain in particular, high standard deviation serves as an indicator of geometric jitter, misalignment between volumes.
The jitter metric $\mathrm{jit}$ is therefore defined for a diffusion-weighted image $I$ and a $b$-value $b$ as:

\begin{equation}
    \mathrm{jit_{map}}(I,b) = \frac{1}{s_0}\ \underset{v\in \mathcal{V}_b}{\operatorname{std}}\left(I_v\right)\ ,\quad \mathrm{jit}(I,b) = \underset{\boldsymbol{x}\in \mathcal{X}_\mathrm{rim}}{\operatorname{mean}}\left(\mathrm{jit_{map}}(I,b)(\boldsymbol{x})  \right)
\end{equation}

\begin{figure}[!h]
    \centering
    \begin{tabular}{c}
        a) \includegraphics[width=0.8\linewidth]{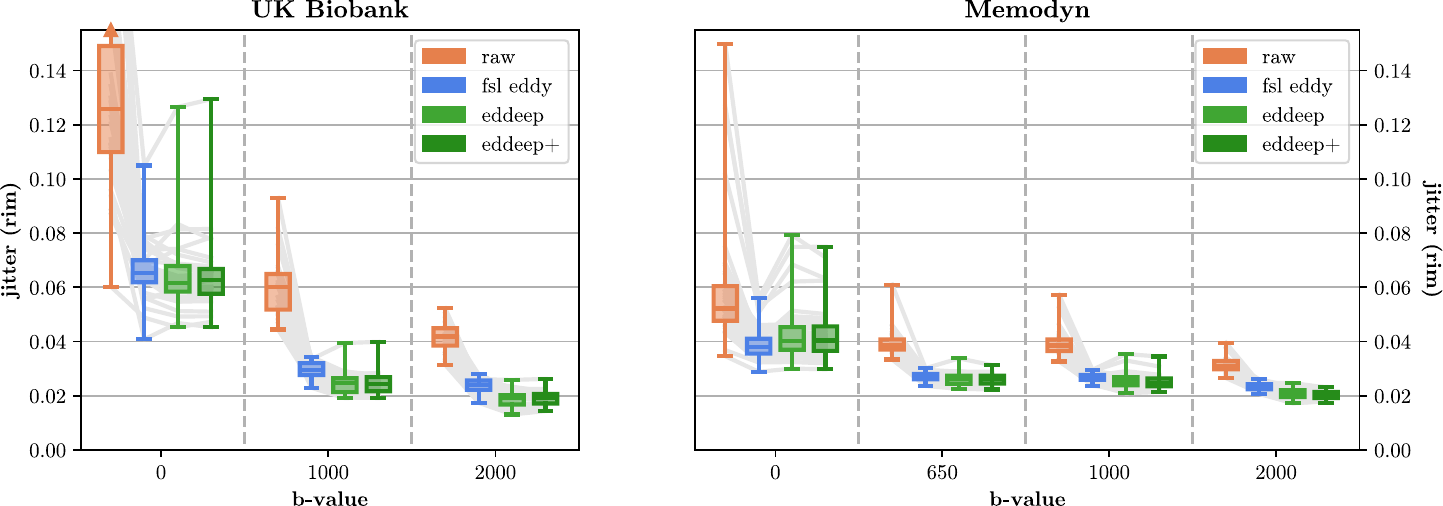}\vspace{0.5cm}\\
        b) \includegraphics[width=0.9\linewidth]{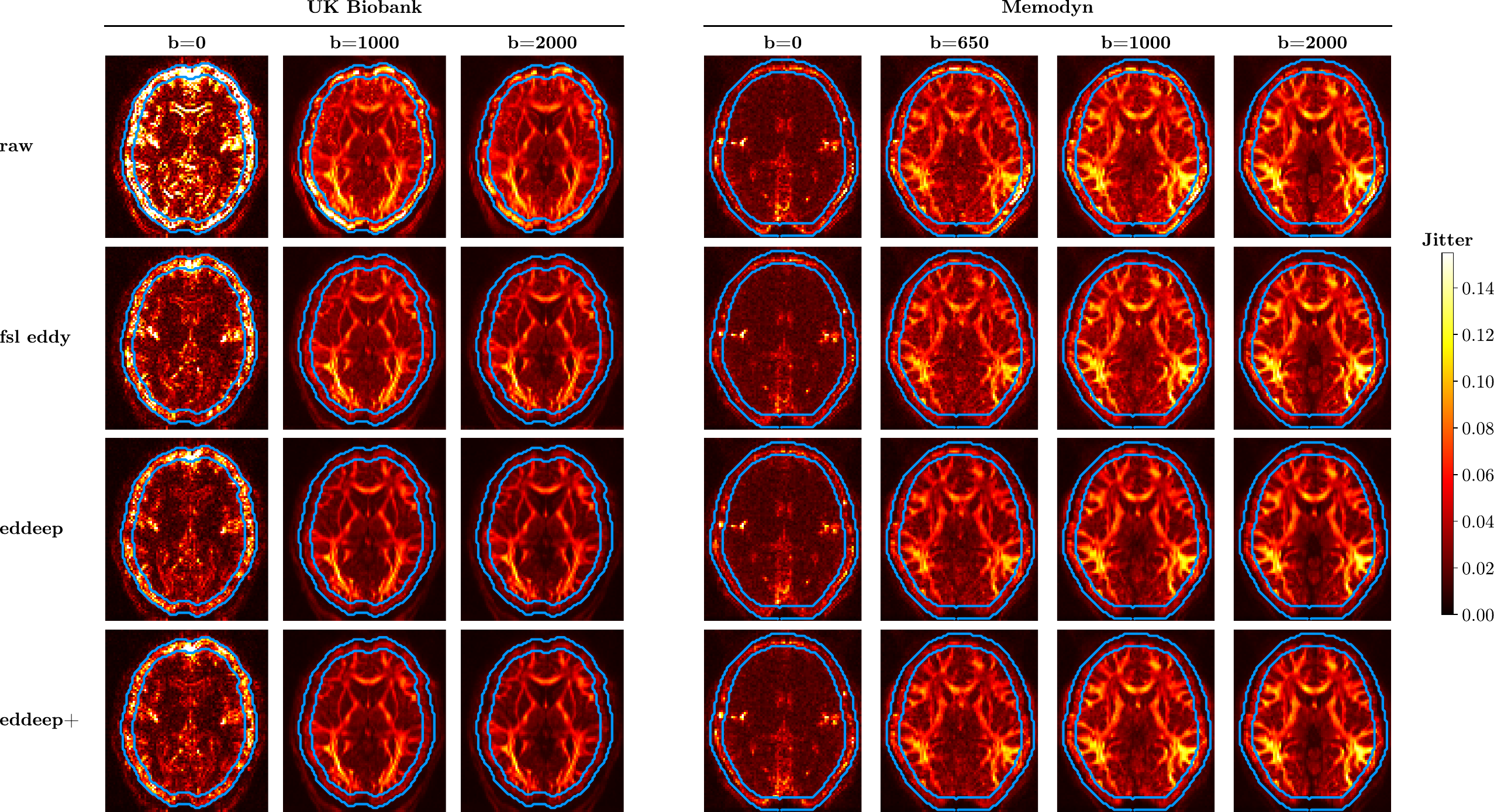}
    \end{tabular}
    \caption{\textbf{Jitter.} a)~Boxplots of $\mathrm{jit}(I,b)$ per $b$-value for uncorrected data, FSL~Eddy, and \emph{Eddeep} variants; lower is better. b)~Voxel-wise map $\mathrm{jit_{map}}$ on a representative subject: the brain mask is outlined in blue; the rim region $\mathcal{X}_\mathrm{rim}$ used to compute the scalar metric is visible at the cortical boundary. Both \emph{Eddeep} and FSL~Eddy strongly reduce jitter at the brain rim, reaching comparable levels on both datasets.}
    \label{jitter}
\end{figure}

\textbf{Results.} Figure~\ref{jitter} shows that the \emph{Eddeep} variants and FSL~Eddy drastically reduce jitter compared with the uncorrected data on both datasets, all reaching overall similar levels. 
At non-zero $b$-values, the \emph{Eddeep} variants achieve significantly lower jitter than FSL~Eddy on both datasets, whereas FSL~Eddy performs better at $b=0$ on Memodyn (Supplementary Material).

\subsubsection{Jitter across translated corrected volumes}
\label{eval_transjit}

We compute the same jitter metric, but this time on the data translated after correction. After translation, the contrast differences are largely removed, so the remaining variability more directly reflects residual geometric misalignment and the metric can be evaluated on the whole brain, not just the rim.
With contrast normalised in this way, higher values are expected at higher $b$-values, which generally induce stronger eddy-current distortions.
To isolate registration quality from translation quality, we use an \emph{oracle} translator. It follows the same training strategy as \texttt{trans+}, but is trained directly on the subjects of the test set. 
Denoting $\tilde{I}$ the translated version of $I$, the metric is computed over the full brain mask $\mathcal{X}_\mathrm{brain}$:

\begin{equation}
    \mathrm{\widetilde{jit}_{map}}(I,b) = \frac{1}{s_0}\ \underset{v\in \mathcal{V}_b}{\operatorname{std}}\left(\tilde{I}_v\right)\ ,\quad \mathrm{\widetilde{jit}}(I,b) = \underset{\boldsymbol{x}\in \mathcal{X}_\mathrm{brain}}{\operatorname{mean}}\left(\mathrm{\widetilde{jit}_{map}}(I,b)(\boldsymbol{x})  \right)
\end{equation}

\begin{figure}[!h]
    \centering
    \begin{tabular}{c}
        a) \includegraphics[width=0.8\linewidth]{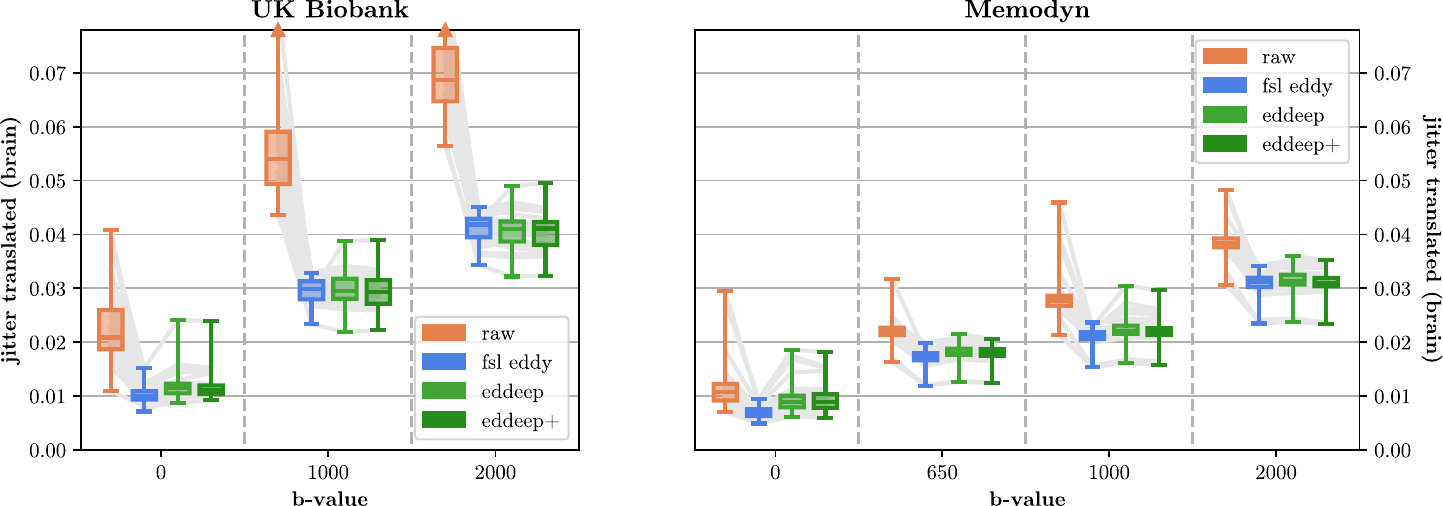}\vspace{0.5cm}\\
        b) \includegraphics[width=0.9\linewidth]{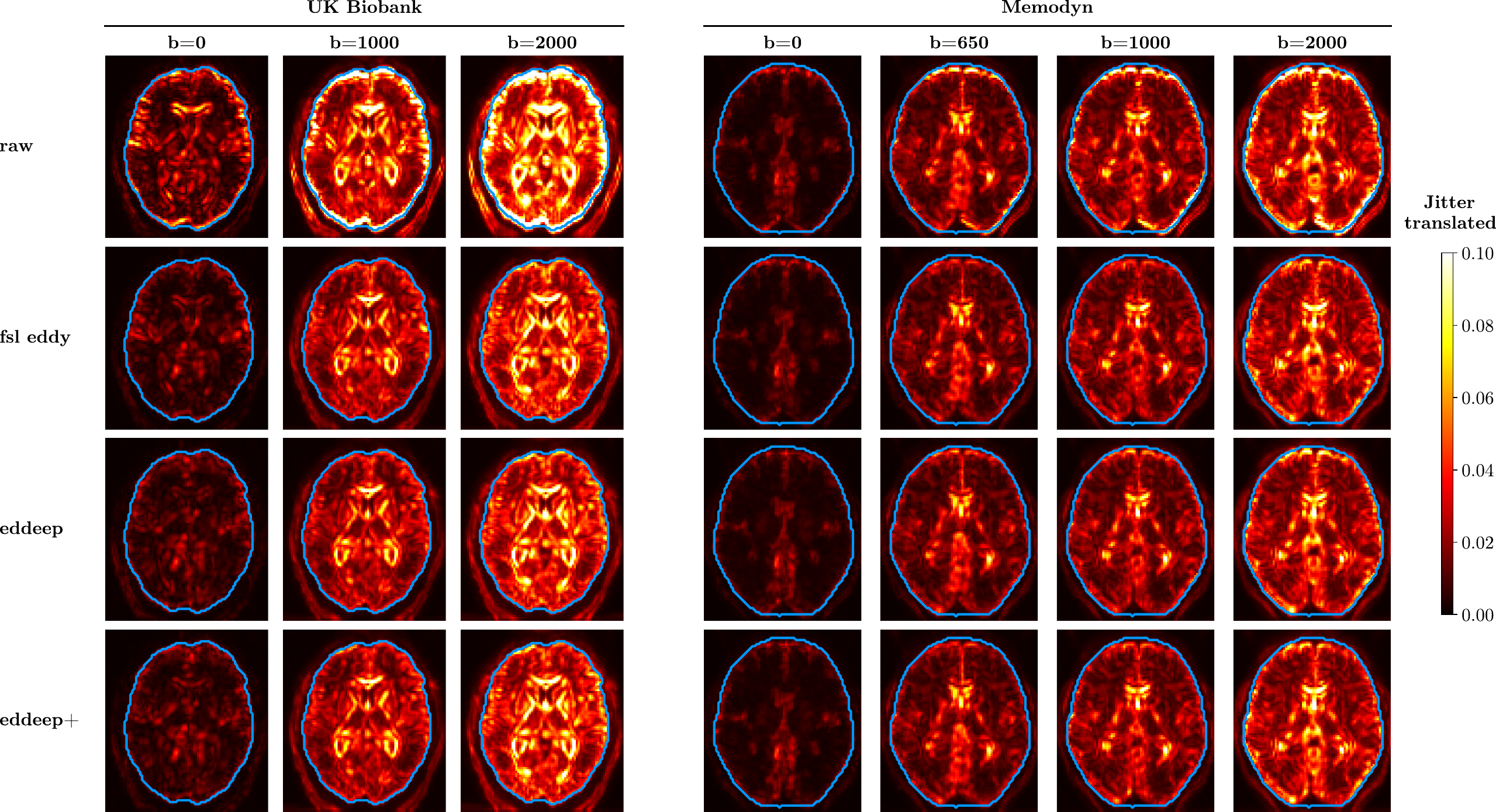}
    \end{tabular}
    \caption{\textbf{Jitter on translated.} a)~Boxplots of $\widetilde{\mathrm{jit}}(I,b)$ per $b$-value for uncorrected data, FSL~Eddy, and \emph{Eddeep} variants; lower is better. b)~Voxel-wise map $\widetilde{\mathrm{jit}}_\mathrm{map}$ on a representative subject after oracle translation, showing residual misalignment across the whole brain rather than only at the rim. After translation, residual misalignment over the whole brain is strongly reduced and comparable for all correction methods.}
    \label{trans_jitter_map}
\end{figure}

\textbf{Results.} Figure~\ref{trans_jitter_map} shows that the \emph{Eddeep} variants and FSL~Eddy drastically reduce jitter compared with the uncorrected data on both datasets. 
FSL~Eddy achieves significantly lower values at $b=0$ on both datasets, and by a small margin at the other shells on Memodyn, while all methods are close at non-zero $b$-values on UK Biobank (Supplementary Material).

\subsubsection{DKI fractional anisotropy halo and model residuals}
\label{eval_dki}

When diffusion-weighted volumes are geometrically misaligned, fitting a diffusion model treats signals from slightly different anatomical locations as if they originated from the same voxel.
This spatial inconsistency manifests particularly at the brain boundary: the cortical rim shifts its position across volumes, so the signals observed at a given boundary voxel alternate between brain tissue and surrounding CSF or background.
When a model is fitted with these signals, the result is an apparent anisotropy signal at the rim that is purely artifactual, the well-known fractional anisotropy (FA) halo~\citep{jones2010}.
We use diffusion kurtosis imaging (DKI)~\citep{jensen2005} rather than DTI because our protocol includes shells up to $b=2000$, for which the Gaussian diffusion assumption of DTI is no longer appropriate; DKI provides a more accurate model and therefore a more sensitive residual metric.

\textbf{FA halo.} The mean FA in the rim region provides a direct readout of boundary inconsistency:
\begin{equation}
    \mathrm{FA}_\mathrm{halo}(I) = \underset{\boldsymbol{x}\in\mathcal{X}_\mathrm{rim}}{\operatorname{mean}}\left(\operatorname{FA}(I)(\boldsymbol{x})\right)
\end{equation}
Lower values indicate better-aligned data; well-corrected data should show low, near-CSF FA values at the rim.

\textbf{DKI residuals.} Beyond the boundary, model residuals over the whole brain provide a complementary measure of data consistency.
Misaligned volumes break the smooth spatial structure assumed by any diffusion model, inflating the mean absolute difference between the observed signal and the DKI fit:
\begin{equation}
\mathrm{res_{map}}(I) = \frac{1}{s_0}\ \underset{v\in \mathcal{V}}{\operatorname{mean}}\left|I_v - \mathrm{DKI}(I_v)\right|\ ,\quad \mathrm{res}(I) = \underset{\boldsymbol{x}\in\mathcal{X}_\mathrm{brain}}{\operatorname{mean}}\left(\mathrm{res_{map}}(I)(\boldsymbol{x})\right)
\end{equation}
where $\mathrm{DKI}(I_v)$ is the signal predicted by the DKI model fitted to $I$.
Lower residuals indicate data that is more consistent with the smooth diffusion model, i.e. better corrected.

\begin{figure}[!h]
    \centering
    \includegraphics[width=\linewidth]{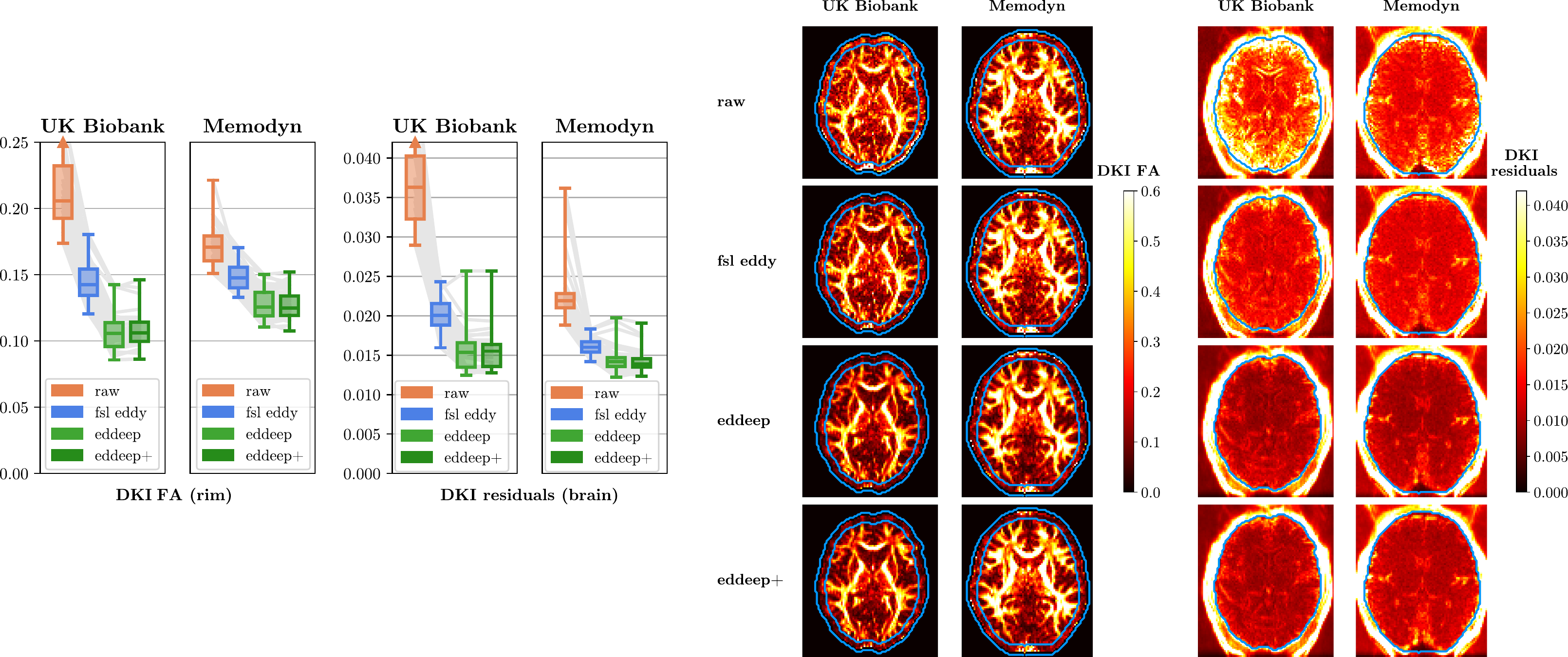}
    \caption{\textbf{DKI-based metrics.} Left: boxplots of FA halo $\mathrm{FA_{halo}}$ for uncorrected data, FSL~Eddy, and \emph{Eddeep} variants; lower is better. Right: boxplots of DKI residuals $\mathrm{res}$; lower is better. Spatial maps of $\mathrm{res_{map}}$ are shown for a representative subject.    
    Correction reduces both the artefactual FA halo at the rim and the DKI residuals, for all correction methods.}
    \label{dki}
\end{figure}

\textbf{Results.} Figure~\ref{dki} shows that the \emph{Eddeep} variants and FSL~Eddy both reduce the FA halo and the DKI residuals compared with the uncorrected data on both datasets. 
For both DKI-based metrics, the \emph{Eddeep} variants achieve significantly lower values than FSL~Eddy on both datasets (Supplementary Material). Although we matched the correction settings as closely as possible, with both FSL~Eddy and the \emph{Eddeep} variants using Jacobian intensity modulation and trilinear interpolation, these metrics are sensitive to implementation differences beyond residual misalignment.

\subsubsection{Signal regularity}
\label{eval_rough}

The diffusion-weighted signal in each voxel is expected to have some regularity in q-space. In particular, for a given $b$-value, a small angular change in gradient direction should not produce abrupt signal variations. In the presence of between-volume misalignment, however, each diffusion direction samples slightly different tissue locations, breaking this regular angular dependence and leading to irregular signal fluctuations across neighbouring directions. The following metric $\mathrm{irreg}$ quantifies this signal irregularity with respect to the gradient direction in a non-parametric fashion, without assuming any signal model:
\begin{equation}
    \mathrm{irreg}(I,b) = \frac{1}{s_0}\ \underset{v\in\mathcal{V}_b}{\operatorname{mean}}\left(\underset{w\in\mathcal{N}_v}{\operatorname{mean}}\left(\frac{\underset{\boldsymbol{x}\in \mathcal{X}_\mathrm{brain}}{\operatorname{mean}}\left(|I_v(\boldsymbol{x})-I_w(\boldsymbol{x})|\right)}{\theta(\boldsymbol{g}_v,\boldsymbol{g}_w)}\right)\right)
\end{equation}
where $\boldsymbol{g}$ are the gradient directions, $\theta$ is the angular distance: $\theta(\boldsymbol{g}_v,\boldsymbol{g}_w)=\arccos(|\boldsymbol{g}_v\cdot \boldsymbol{g}_w|)$, and $\mathcal{N}_v$ is the set of indices of the $k$-nearest angular neighbours (according to $\theta$) of $\boldsymbol{g}_v$; we use $k=3$ in practice. For $b=0$, where no diffusion gradient is applied, volumes are expected to be identical in q-space, so we set $\theta=1$ to measure plain signal difference between $b=0$ volumes.

\begin{figure}[!h]
    \centering
    \begin{tabular}{c}
        a) \includegraphics[width=0.8\linewidth]{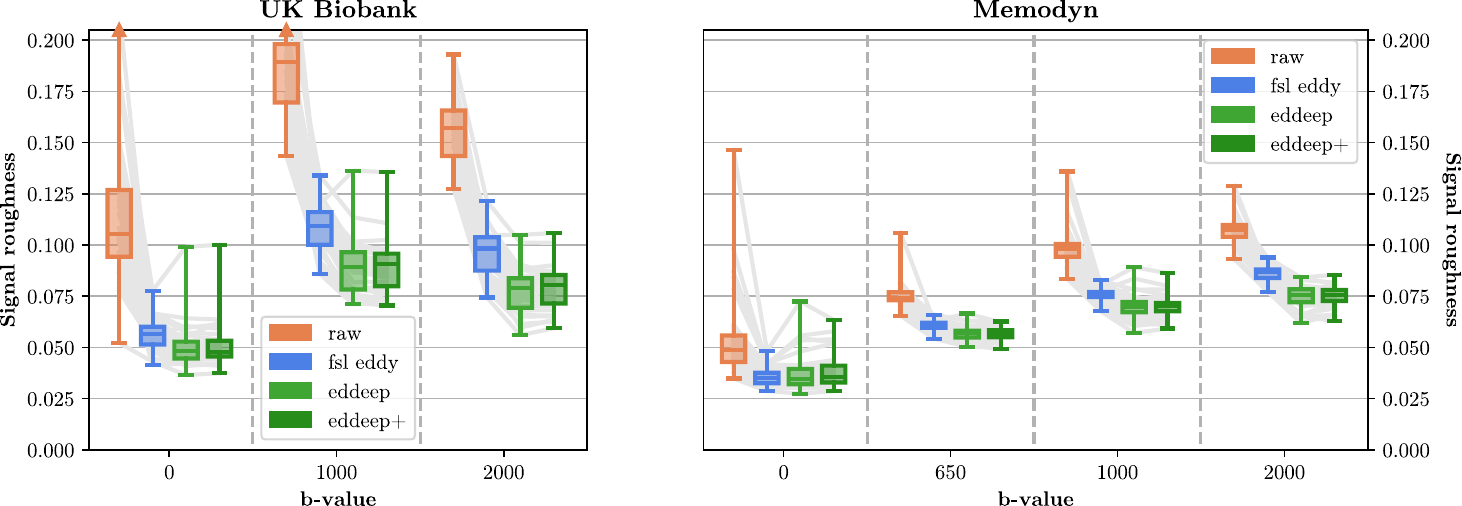}\vspace{0.5cm}\\
        b) \includegraphics[width=0.9\linewidth]{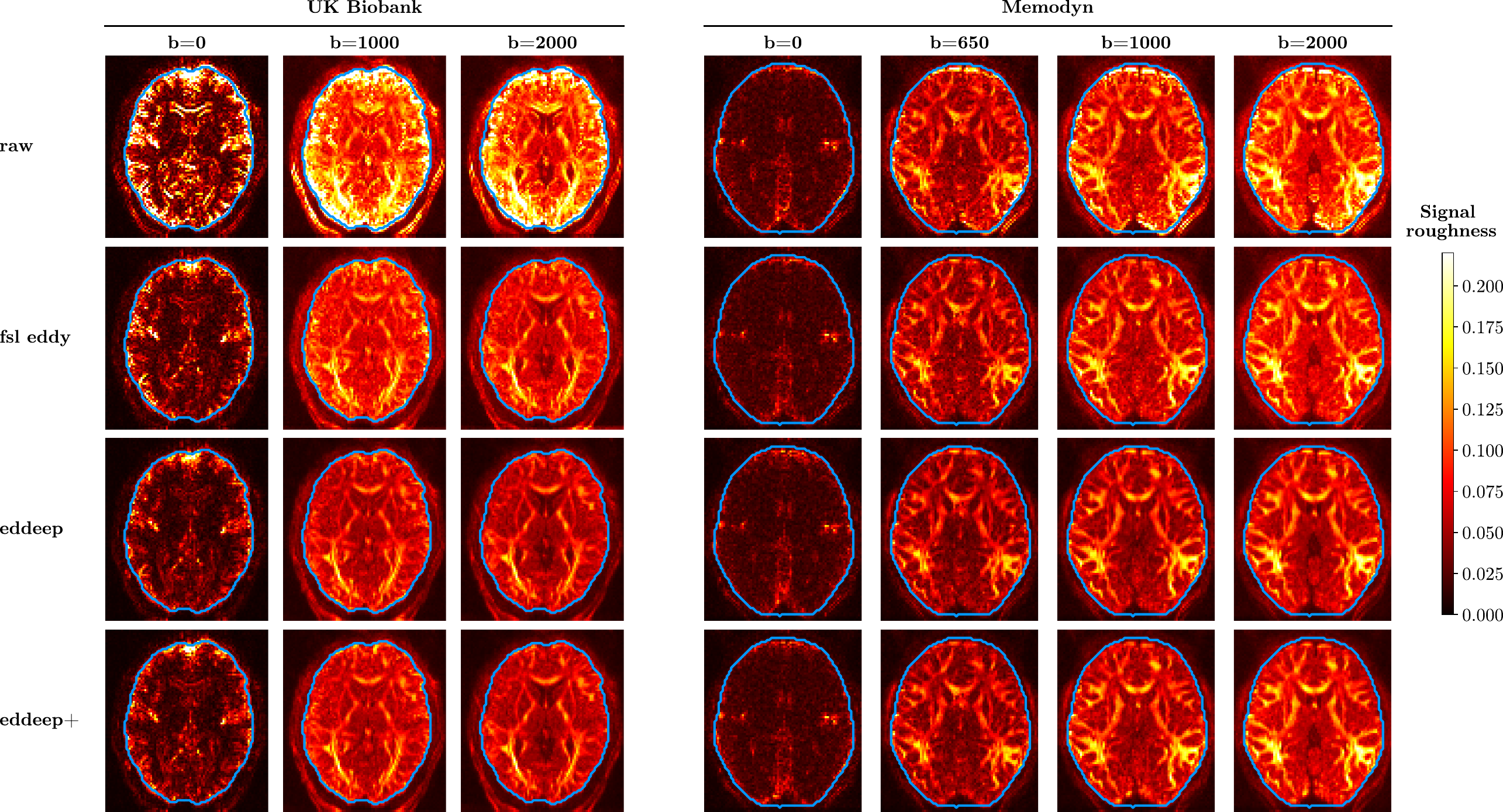}
    \end{tabular}
    \caption{\textbf{Signal irregularity metric.} a)~Boxplots per $b$-value for uncorrected data, FSL~Eddy, and \emph{Eddeep} variants; lower is better. b)~Voxel-wise map of angular irregularity on a representative subject, highlighting regions where between-volume misalignment disrupts the expected regularity of the signal in q-space. Correction restores the regularity of the signal in q-space, with lower irregularity for \emph{Eddeep} at non-zero $b$-values.}
    \label{sig_rough}
\end{figure}

\textbf{Results.} Figure~\ref{sig_rough} shows that the \emph{Eddeep} variants and FSL~Eddy both reduce signal irregularity compared with the uncorrected data on both datasets.
At non-zero $b$-values, the \emph{Eddeep} variants achieve significantly lower irregularity than FSL~Eddy on both datasets, while FSL~Eddy is slightly better at $b=0$ on Memodyn (Supplementary Material).

\subsubsection{Mutual information}
\label{eval_mi}

As discussed in the introduction (Section~\ref{intro}), mutual information (MI) is not reliable as a registration cost function for eddy distortion correction. Nonetheless, it is informative at evaluation time. MI measures the statistical dependence between pairs of volumes, which is influenced by how well they are spatially aligned. We use normalised mutual information (NMI), as defined in~\cite{studholme1999}, averaged over all volume pairs for a given shell combination $(b_1,b_2)$, which provides a more robust estimate than a single pairwise comparison.

\begin{equation}
\mathrm{MI_{pair}}(I, b_1, b_2) = \underset{v_1 \in \mathcal{V}_{b_1},\; v_2 \in \mathcal{V}_{b_2}}{\operatorname{mean}} \mathrm{NMI}\left(I_{v_1},\, I_{v_2}\right)
\end{equation}
NMI lies in $[1, 2]$: a value of 1 indicates statistical independence between the two volumes, and 2 indicates perfect correlation; higher is better.

\begin{figure}[!h]
    \centering
    \includegraphics[width=\linewidth]{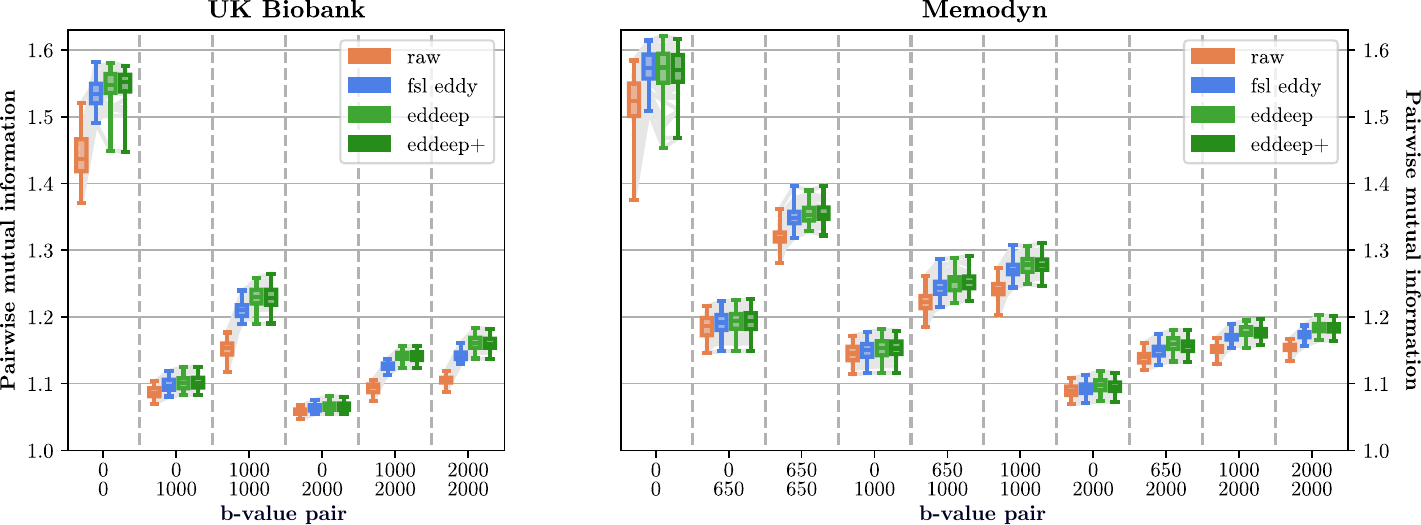}
    \caption{\textbf{Normalised mutual information}. Pairwise, per $b$-value pair for uncorrected data, FSL~Eddy, and \emph{Eddeep} variants; higher is better. Correction increases the similarity between volumes across all shell pairs, with comparable values for all correction methods.}
    \label{mi_pair}
\end{figure}

\textbf{Results.} Figure~\ref{mi_pair} shows that the \emph{Eddeep} variants and FSL~Eddy both increase pairwise NMI compared with the uncorrected data across all shell combinations and on both datasets. 
The \emph{Eddeep} variants achieve significantly higher NMI than FSL~Eddy for almost all shell pairs, although the differences are small (Supplementary Material).

\subsubsection{Training-run variability}
\label{eval_seeds}

The results above are obtained with a single trained model per variant, and therefore characterise variability across subjects but not across training runs. To assess the latter, we retrained the registration model of \texttt{eddeep+} with five additional random seeds, using identical settings and a fixed \texttt{trans+} translation model. The model used throughout the evaluation above corresponds to seed~1.

\begin{figure}[!h]
    \centering
    \includegraphics[width=0.9\linewidth]{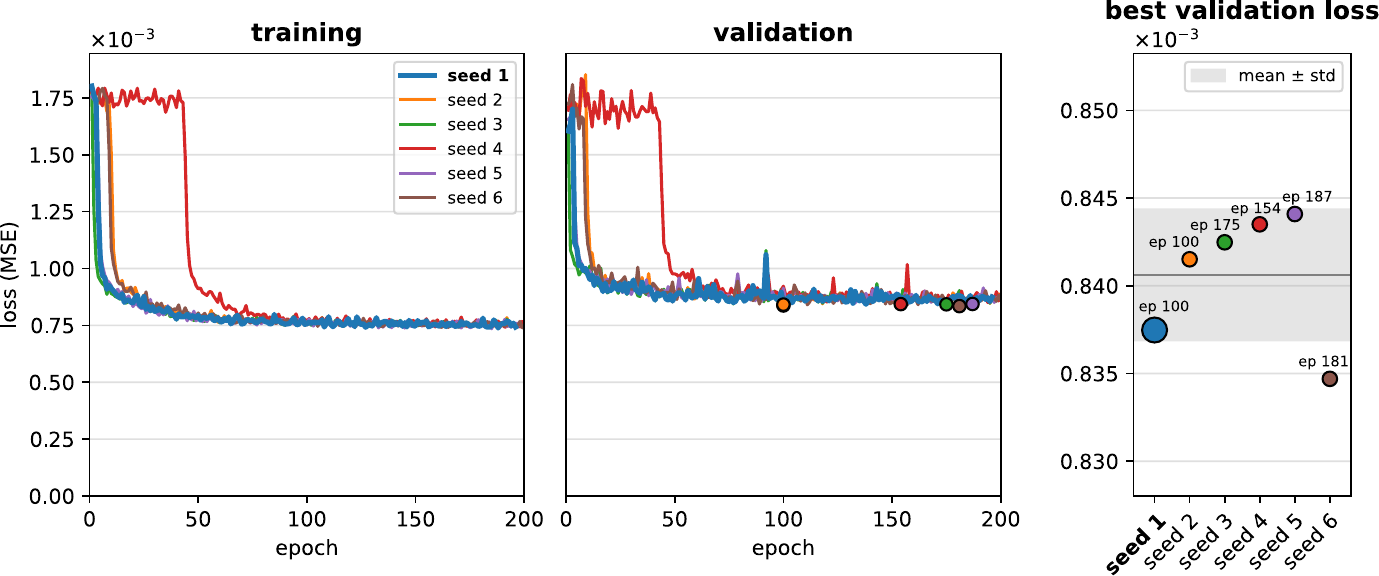}
    \caption{\textbf{Training-run variability.} Training and validation loss (MSE) of the registration model for six random seeds; seed~1 is the model used throughout this work. Right: best validation loss per seed, annotated with the epoch at which it is reached; the grey band shows the mean $\pm$ std. All runs converge to the same plateau.}
    \label{seeds}
\end{figure}

\textbf{Results.} Figure~\ref{seeds} shows that the loss first stays at its initial level, corresponding to a near-identity transformation, for a variable number of epochs before dropping abruptly: 4 to 12 epochs for most seeds, but 47 for one of them. 
We hypothesise that this plateau results from the near-identity initialisation~\citep{saxe2014}, combined with a loss that constrains only a few global parameters, and only indirectly through the warped image.
Once past this plateau, all six runs converge to the same level, with a coefficient of variation of 0.44\,\% on the best validation loss. Seed variability therefore mainly affects the speed of convergence, not the final model.

\subsection{Processing time}
\label{proctime}

We compared the wall-clock time required to correct a single acquisition with \emph{Eddeep} (\texttt{eddeep+}, inference) and with the FSL~Eddy baseline; their configurations are detailed in Sections~\ref{eddepmodels} and~\ref{baseline} respectively. To reflect realistic usage, we considered two hardware platforms and, on each, ran every method with the best computational backend available to it on that platform:

\begin{itemize}
    \item \emph{Workstation.} Linux workstation (Rocky Linux~9.8; 8-core CPU, NVIDIA GeForce RTX~3090 GPU with 24\,GB VRAM, driver 610.43.02, CUDA~12.2). Both methods ran on GPU: \emph{Eddeep} via TensorFlow~2.18.0 (cuDNN~9.3), and FSL~Eddy via its CUDA implementation (\texttt{eddy\_cuda10.2}), whose CPU-side computations are single-threaded.

    \item \emph{Laptop.} Apple M3 MacBook Pro (Mac15,3; 8-core CPU, 10-core GPU, 24\,GB unified memory; macOS~14.3, Metal~3). \emph{Eddeep} ran on GPU via TensorFlow~2.16.1 with the tensorflow-metal~1.2.0 plugin. FSL~Eddy has no GPU implementation for this platform, its CUDA version requiring an NVIDIA GPU; its best available option is therefore the multi-threaded CPU implementation (\texttt{eddy\_openmp}), run across all CPU cores.

\end{itemize}

\textbf{Results.} Results are reported in Table~\ref{tab_proctime}. 
On the workstation, where both methods run on the same GPU and the comparison is therefore matched in backend, \emph{Eddeep} corrected a full acquisition in about \SI{21}{\second}, against \SIrange{2}{3}{\minute} for FSL~Eddy, a 6 to 8-fold reduction. On the laptop, where the comparison is platform-specific but reflects the runtime a user would experience on such hardware, \emph{Eddeep} took a little over \SI{2}{\minute}, against more than one hour for FSL~Eddy, a more than 30-fold reduction. 
The gap widens on the laptop because \emph{Eddeep}, built on a standard deep-learning framework, can use GPU acceleration on any hardware supported by that framework, whereas FSL~Eddy's GPU acceleration is restricted to NVIDIA hardware. The reported times reflect the end-to-end runtime of each tool: each FSL~Eddy run also performs computations that do not affect the resampled output, such as outlier detection and QC-related estimates, which remain active even when outlier replacement is disabled.

\begin{table}[!h]
    \centering
    \begin{tabular}{l l c cc cc c}
    \toprule
    & & & \multicolumn{2}{c}{\emph{Eddeep}} & \multicolumn{2}{c}{FSL Eddy} & \\
    \cmidrule(lr){4-5} \cmidrule(lr){6-7}
    Platform & Dataset & $n$ & mean & std & mean & std & speed-up \\
    \midrule
    \multirow{2}{*}{Workstation}
      & UKB     & 105 & \SI{21.1}{\second} & \SI{1.6}{\second}
                      & \SI{2}{\minute}\,\SI{14}{\second} & \SI{7.6}{\second} & $\times 6.4$ \\
      & Memodyn & 118 & \SI{21.9}{\second} & \SI{1.8}{\second}
                      & \SI{2}{\minute}\,\SI{52}{\second} & \SI{4.3}{\second} & $\times 7.8$ \\
    \midrule
    \multirow{2}{*}{Laptop}
      & UKB     & 105 & \SI{2}{\minute}\,\SI{15}{\second} & \SI{3}{\second}
                      & \SI{1}{\hour}\,\SI{11}{\minute}\,\SI{26}{\second} & \SI{1}{\minute}\,\SI{5}{\second} & $\times 31.7$ \\
      & Memodyn & 118 & \SI{2}{\minute}\,\SI{26}{\second} & \SI{3}{\second}
                      & \SI{1}{\hour}\,\SI{21}{\minute}\,\SI{11}{\second} & \SI{1}{\minute}\,\SI{47}{\second} & $\times 33.4$ \\
    \bottomrule
    \end{tabular}
    
    \caption{\textbf{Processing time}. Wall-clock time for the eddy-current distortion correction of one acquisition of $n$ volumes with \emph{Eddeep} (inference) and FSL~Eddy. Each run with the best backend available on the platform. \emph{Workstation}: both on an NVIDIA RTX~3090 GPU (FSL~Eddy: \texttt{eddy\_cuda10.2}). \emph{Laptop}: Apple M3 MacBook Pro, \emph{Eddeep} on GPU (Metal) and FSL~Eddy on CPU (\texttt{eddy\_openmp}, multi-threaded).}
    \label{tab_proctime}
\end{table}

\section{Conclusion and Discussion}

In this work, we introduced \emph{Eddeep}, a two-stage deep-learning framework for fast eddy-current distortion correction in diffusion MRI. By combining image translation for contrast standardisation with a registration model inspired by the physics of EPI acquisition, \emph{Eddeep} directly maps each diffusion-weighted volume to an undistorted geometry referenced to $b=0$. Unlike conventional approaches, it does not rely on an iterative prediction-correction scheme, but instead seeks the transformation to the reference space in a single forward pass. At inference time, it also makes no assumption about the sampling of q-space. In particular, since each volume is corrected independently of the others, \emph{Eddeep} can be applied without modification to acquisitions with only a few diffusion directions, such as clinical protocols, where prediction-based methods have little data from which to build their prediction. Across both in-domain and out-of-domain datasets, \emph{Eddeep} achieved correction quality comparable to FSL~Eddy, with an advantage on the metrics most directly tied to residual distortion and a disadvantage at $b=0$, while offering faster inference. Beyond the models themselves, we also provide task-specific augmentation strategies and an openly available implementation with documentation, which we hope will facilitate adoption and further development.

For advanced diffusion MRI to become practical in routine clinical settings, the entire processing pipeline, from raw image acquisition through distortion correction to the estimation of microstructural parameter maps, must operate within a clinically acceptable time frame. Computationally expensive components based on slow iterative optimisation are difficult to reconcile with such streamlined workflows. This motivates the development of machine-learning approaches capable of rapid inference and aligns with the recent support by MRI vendors for the inline deployment of trained neural networks for image manipulation tasks. In this context, \emph{Eddeep} is complementary to existing deep-learning methods for susceptibility distortion correction~\citep{legouhy2022,duong2020,zaid2024} and microstructural parameter estimation~\citep{barbieri2020,golkov2016,sen2024,kerkela2026,epstein2024,sen2026}. More broadly, recent proof-of-concept work has demonstrated the feasibility of inline quantitative MRI~\citep{rot2026}.

The present study has several limitations. First, as no ground-truth eddy distortions are available in real data, evaluation must rely on indirect but complementary metrics, each reflecting only one aspect of correction quality. Second, training the translation model requires previously-corrected data generated with an external tool, which introduces a dependence on an existing correction method to build the training set. Failed corrections can be discarded by quality-checking the training data, but residual errors can impact the quality of the translator.

Third, the metrics reported here were computed with the acquired $b$-vectors. Strictly, when a rigid rotation is estimated and applied to a diffusion-weighted volume, the associated $b$-vector should be rotated with it, otherwise the corrected image and its $b$-vector no longer describe the same physical direction~\citep{leemans2009}. In our experiments, this has a negligible effect: the estimated between-volume rotations are sub-degree (median $0.36^\circ$ and $0.27^\circ$, 95th percentile $0.96^\circ$ and $0.65^\circ$, on UKB and Memodyn respectively), and recomputing the $b$-vector-dependent metrics with rotated $b$-vectors changes the DKI residuals by less than $0.01\%$ on average ($0.05\%$ at most for a single subject) and the signal regularity by at most $0.1\%$. These differences are orders of magnitude smaller than those between the methods compared here. Rotations can be considerably larger in other settings, however, such as longer acquisitions or paediatric and patient populations, where this correction becomes important. For that reason, the correction script released with \emph{Eddeep} provides an option to output the rotated $b$-vectors.

Fourth, \emph{Eddeep} does not yet provide some advanced features of recent versions of FSL~Eddy, such as correction for within-volume motion~\citep{andersson2017}, which can occur during the acquisition of a single volume, and outlier replacement~\citep{andersson2016b}, whereby slices with signal strongly inconsistent with the model prediction, typically due to motion-induced dropout, are identified and replaced. In the present work, however, we deliberately focus on the eddy-current distortion correction component itself.

Finally, and most importantly, the generalisability of \emph{Eddeep} beyond the tested domain remains to be established. \emph{Eddeep} generalised well to an out-of-domain dataset, Memodyn, acquired on a different scanner model with a different protocol. This protocol includes a $b=650$\,s/mm² shell, absent from training, which was corrected on par with FSL~Eddy (Fig.~\ref{jitter}), showing that the translator interpolates reliably to unseen $b$-values. Both datasets nonetheless share several characteristics: 3T Siemens scanners, 2\,mm isotropic voxels, antero-posterior phase encoding, $b\leq2000$\,s/mm² and healthy adult brains. Since the registration model only sees translated images, generalisation beyond this domain mostly rests on the translator. Extrapolation to much higher $b$-values is more uncertain, as contrasts depart further from those seen in training and SNR drops. When tissue boundaries are buried in noise, the translation is driven only by the remaining white matter signal and the learned prior, which is likely to degrade its output. Populations whose tissue contrast differs markedly from that of adults, such as neonates and children, as well as pathological brains, other organs, phantoms, or other phase-encoding directions, would likely require retraining. Other vendors, field strengths and larger distortion amplitudes also remain to be tested.

Several directions for future work follow naturally from these limitations. The main objective of future iterations of \emph{Eddeep} is to extend its generalisability, by training on more diverse data and strengthening augmentation, for example through domain randomisation~\citep{billot2023} or lesion synthesis. Although the quadratic distortion model used here provides a useful compromise between flexibility and parsimony, polynomial terms can grow rapidly away from the image centre and are not necessarily the most faithful representation of the underlying distortion physics. A low-frequency deformation field may provide a more realistic alternative while preserving regularity.

We also plan to investigate the synthesis of diffusion-weighted images from anatomical images~\citep{ren2021}, as a way to obtain undistorted references without external correction. Conditioning the translator on the $b$-value and $b$-vector, so as to synthesise the expected contrast of each volume, would in turn provide a natural basis for outlier detection and replacement, and slice-to-volume correction. The registration model would, however, then face inputs of varying contrast, and share with the translator the burden of generalising across contrasts, which the present design places on the translator alone.
\section*{Data and Code Availability}

The code for \emph{Eddeep} is available open-source, with documentation and examples, at \url{https://github.com/CIG-UCL/eddeep}. The trained models used for all results reported in this work are released in the \texttt{models/} directory of that repository, for both the \texttt{eddeep} and \texttt{eddeep+} variants.

\section*{Author Contributions}

A.L.: Conceptualisation, Methodology, Validation, Software (lead dev.), Writing (original draft).\\
R.C.: Methodology (through discussion), Writing (review and editing).\\
Y.Q.: Resources, Writing (review).\\
W.S.: Resources, Writing (review).\\
P.P.: Resources, Writing (review).\\
H.A.: Funding acquisition, Project administration, Resources, Writing (review).\\
H.Z.: Funding acquisition, Project administration, Resources, Supervision, Conceptualisation, Writing (review and editing).

\section*{Ethics statement}
This study used two existing diffusion MRI datasets. UK Biobank has approval from the North West Multi-centre Research Ethics Committee (MREC) as a Research Tissue Bank, and all participants gave written informed consent. The Memodyn study was approved by the Liège University Hospital Ethics Committee (approval \#2020/138), and all participants provided written informed consent.

\section*{Funding}

H.A. was supported by UKRI Future Leaders Fellowship MR/W011980/1 – QUANTIMA: Quantitative imaging platform for the diagnosis, subtyping, staging and outcome prognosis in dementia.\\
A.L., R.C., H.A., and H.Z. were supported by Innovate UK grant 10036158 – CLAIR: first-in-class non-invasive test for outcome prediction and stratification in dementia.\\
A.L. and H.Z. were supported by National Institute for Health (1R01MH130362).\\
W.S was supported by the Fonds de la Recherche Scientifique (FRS-FNRS., Aspirant Research Fellowship).\\
The study was supported by the FNRS and the Fonds Wetenschappelijk Onderzoek--Vlaanderen (FWO) under the Excellence of Science (EOS) Project (MEMODYN, No. 30446199 to P.P. and H.Z.).\\
This work was supported by the ERC grant (UNFOLD, ERC-2023-SyG n°101118729). Funded by the European Union. Views and opinions expressed are however those of the author(s) only and do not necessarily reflect those of the European Union or the European Research Council Executive Agency. Neither the European Union nor the granting authority can be held responsible for them.\\
For the purpose of open access, the author has applied a CC-BY public copyright licence to any Author Manuscript version arising from this submission. 

\section*{Declaration of Competing Interests}

We declare we do not have conflicts of interest.





\printbibliography

@article{leemans2009,
  title={The B-matrix must be rotated when correcting for subject motion in DTI data},
  author={Leemans, Alexander and Jones, Derek K},
  journal={Magnetic Resonance in Medicine: An Official Journal of the International Society for Magnetic Resonance in Medicine},
  volume={61},
  number={6},
  pages={1336--1349},
  year={2009},
  publisher={Wiley Online Library}
}

@article{blanco2021,
  title={A tutorial on $SE(3)$ transformation parameterizations and on-manifold optimization},
  author={Blanco-Claraco, Jos{\'e} Luis},
  journal={arXiv preprint arXiv:2103.15980},
  year={2021},
  url={https://arxiv.org/abs/2103.15980}
}

@article{zaid2024,
  title={FD-Net: An unsupervised deep forward-distortion model for susceptibility artifact correction in EPI},
  author={Zaid Alkilani, Abdallah and {\c{C}}ukur, Tolga and Saritas, Emine Ulku},
  journal={Magnetic resonance in medicine},
  volume={91},
  number={1},
  pages={280--296},
  year={2024},
  publisher={Wiley Online Library}
}

@article{sen2026,
  title={microTorch: A software package for fast and flexible self-supervised diffusion MRI model fitting},
  author={Sen, Snigdha and Ahmed, Rajib and Arends, Gerrit C and Casali, Nicola and Gomez, Alvaro Planchuelo and Chen, Xiaoxiang and Masramon, Marta and Parker, Christopher S and Palombo, Marco and Tax, Chantal MW and others},
  journal={Journal of Open Research Software},
  volume={14},
  number={1},
  pages={59--59},
  year={2026},
  publisher={Ubiquity Press}
}

@article{alexander1997,
  title={Elimination of eddy current artifacts in diffusion-weighted echo-planar images: the use of bipolar gradients},
  author={Alexander, Andrew L and Tsuruda, Jay S and Parker, Dennis L},
  journal={Magnetic Resonance in Medicine},
  volume={38},
  number={6},
  pages={1016--1021},
  year={1997},
  publisher={Wiley Online Library}
}

@article{barmet2008,
  title={Spatiotemporal magnetic field monitoring for MR},
  author={Barmet, Christoph and Zanche, Nicola De and Pruessmann, Klaas P},
  journal={Magnetic Resonance in Medicine: An Official Journal of the International Society for Magnetic Resonance in Medicine},
  volume={60},
  number={1},
  pages={187--197},
  year={2008},
  publisher={Wiley Online Library}
}

@inproceedings{saxe2014,
  author    = {Saxe, Andrew M. and McClelland, James L. and Ganguli, Surya},
  title     = {Exact solutions to the nonlinear dynamics of learning in deep linear neural networks},
  booktitle = {International Conference on Learning Representations (ICLR)},
  year      = {2014},
  note      = {arXiv:1312.6120}
}

@article{sotiropoulos2013,
  title={Advances in diffusion MRI acquisition and processing in the Human Connectome Project},
  author={Sotiropoulos, Stamatios N and Jbabdi, Saad and Xu, Junqian and Andersson, Jesper L and Moeller, Steen and Auerbach, Edward J and Glasser, Matthew F and Hernandez, Moises and Sapiro, Guillermo and Jenkinson, Mark and others},
  journal={Neuroimage},
  volume={80},
  pages={125--143},
  year={2013},
  publisher={Elsevier}
}

@article{stee2023,
  title={Microstructural dynamics of motor learning and sleep-dependent consolidation: A diffusion imaging study},
  author={Stee, Whitney and Legouhy, Antoine and Guerreri, Michele and Villemonteix, Thomas and Zhang, Hui and Peigneux, Philippe},
  journal={Iscience},
  volume={26},
  number={12},
  year={2023},
  publisher={Elsevier}
}

@article{diederik2014,
  title={Adam: A method for stochastic optimization},
  author={Kingma, Diederik P. and Ba, Jimmy},
  url={https://arxiv.org/abs/1412.6980},
  year={2014}
}

@article{studholme1999,
  title={An overlap invariant entropy measure of 3D medical image alignment},
  author={Studholme, Colin and Hill, Derek LG and Hawkes, David J},
  journal={Pattern recognition},
  volume={32},
  number={1},
  pages={71--86},
  year={1999},
  publisher={Elsevier}
}

@article{alfaro2018,
  author  = {Alfaro-Almagro, Fidel and Jenkinson, Mark and Bangerter, Neal K. and Andersson, Jesper L.R. and Griffanti, Ludovica and Douaud, Gwenaelle and Sotiropoulos, Stamatios N. and Jbabdi, Saad and Hernandez-Fernandez, Moises and Vallee, Emmanuel and Vidaurre, Diego and Webster, Matthew and McCarthy, Paul and Rorden, Christopher and Daducci, Alessandro and Alexander, Daniel C. and Zhang, Hui and Dragonu, Iulius and Matthews, Paul M. and Miller, Karla L. and Smith, Stephen M.},
  title   = {Image processing and Quality Control for the first 10,000 brain imaging datasets from {UK} {Biobank}},
  journal = {NeuroImage},
  year    = {2018},
  volume  = {166},
  pages   = {400--424},
  url     = {https://doi.org/10.1016/j.neuroimage.2017.10.034}
}

@article{jensen2005,
    author  = {Jensen, Jens H. and Helpern, Joseph A. and Ramani, Anita and Lu, Hanzhang and Kaczynski, Kyle},
    title   = {Diffusional kurtosis imaging: The quantification of non-{G}aussian water diffusion by means of magnetic resonance imaging},
    journal = {Magnetic Resonance in Medicine},
    year    = {2005},
    volume  = {53},
    number  = {6},
    pages   = {1432--1440},
    doi     = {10.1002/mrm.20508}
}

@InProceedings{eddeep,
author="Legouhy, Antoine and Callaghan, Ross and Stee, Whitney and Peigneux, Philippe and Azadbakht, Hojjat and Zhang, Hui",
title="Eddeep: Fast Eddy-Current Distortion Correction for Diffusion MRI with Deep Learning",
booktitle="Medical Image Computing and Computer Assisted Intervention -- MICCAI 2024",
year="2024",
pages="152--161",
isbn="978-3-031-72069-7",
url = {https://doi.org/10.1007/978-3-031-72069-7_15}
}

@book{bernstein2004,
  title={Handbook of MRI pulse sequences},
  author={Bernstein, Matt A and King, Kevin F and Zhou, Xiaohong Joe},
  year={2004},
  publisher={Elsevier}
}

@article{ahn1991,
  title={Analysis of the eddy-current induced artifacts and the temporal compensation in nuclear magnetic resonance imaging},
  author={Ahn, CB and Cho, ZH},
  journal={IEEE transactions on medical imaging},
  volume={10},
  number={1},
  pages={47--52},
  year={1991},
  publisher={IEEE}
}

@misc{ren2021,
      title={Q-space Conditioned Translation Networks for Directional Synthesis of Diffusion Weighted Images from Multi-modal Structural MRI}, 
      author={Mengwei Ren and Heejong Kim and Neel Dey and Guido Gerig},
      year={2021},
      eprint={2106.13188},
      archivePrefix={arXiv},
      primaryClass={eess.IV}
}

@article{miller2016,
  title={Multimodal population brain imaging in the UK Biobank prospective epidemiological study},
  author={Miller, Karla L and Alfaro-Almagro, Fidel and Bangerter, Neal K and Thomas, David L and Yacoub, Essa and Xu, Junqian and Bartsch, Andreas J and Jbabdi, Saad and Sotiropoulos, Stamatios N and Andersson, Jesper LR and others},
  journal={Nature neuroscience},
  volume={19},
  number={11},
  pages={1523--1536},
  year={2016},
  publisher={Nature Publishing Group}
}

@article{haselgrove1996,
author = {Haselgrove, John C. and Moore, James R.},
title = {Correction for distortion of echo-planar images used to calculate the apparent diffusion coefficient},
journal = {Magnetic Resonance in Medicine},
volume = {36},
number = {6},
pages = {960-964},
year = {1996},
url = {https://doi.org/10.1002/mrm.1910360620},
}

@article{rot2026,
author = {Rot, Samuel and Dragonu, Iulius and Triantafyllou, Christina and Grech-Sollars, Matthew and Papadaki, Anastasia and Mancini, Laura and Wastling, Stephen and Steeden, Jennifer and Thornton, John S. and Yousry, Tarek and Gandini Wheeler-Kingshott, Claudia A. M. and Thomas, David L. and Alexander, Daniel C. and Zhang, Hui},
title = {Real-Time, Inline Quantitative MRI Enabled by Scanner-Integrated Machine Learning: A Proof of Principle With NODDI},
journal = {Magnetic Resonance in Medicine},
year = 2026,
url = {https://doi.org/10.1002/mrm.70388}
}

@article{jezzard1999,
  title={Sources of distortion in functional MRI data},
  author={Jezzard, Peter and Clare, Stuart},
  journal={Human brain mapping},
  volume={8},
  number={2-3},
  pages={80--85},
  year={1999},
  publisher={Wiley Online Library}
}

@article{jones2010,
author = {Jones, Derek K. and Cercignani, Mara},
title = {Twenty-five pitfalls in the analysis of diffusion MRI data},
journal = {NMR in Biomedicine},
volume = {23},
number = {7},
pages = {803-820},
url = {https://doi.org/10.1002/nbm.1543},
year = {2010}
}

@article{barbieri2020,
author = {Barbieri, Sebastiano and Gurney-Champion, Oliver J. and Klaassen, Remy and Thoeny, Harriet C.},
title = {Deep learning how to fit an intravoxel incoherent motion model to diffusion-weighted MRI},
journal = {Magnetic Resonance in Medicine},
volume = {83},
number = {1},
pages = {312-321},
url = {https://doi.org/10.1002/mrm.27910},
year = {2020}
}

@article{golkov2016,
  author={Golkov, Vladimir and Dosovitskiy, Alexey and Sperl, Jonathan I. and Menzel, Marion I. and Czisch, Michael and Sämann, Philipp and Brox, Thomas and Cremers, Daniel},
  journal={IEEE Transactions on Medical Imaging}, 
  title={q-Space Deep Learning: Twelve-Fold Shorter and Model-Free Diffusion MRI Scans}, 
  year={2016},
  volume={35},
  number={5},
  pages={1344-1351},
  url={10.1109/TMI.2016.2551324}}

@article{sen2024,
author = {Sen, Snigdha and Singh, Saurabh and Pye, Hayley and Moore, Caroline M. and Whitaker, Hayley C. and Punwani, Shonit and Atkinson, David and Panagiotaki, Eleftheria and Slator, Paddy J.},
title = {ssVERDICT: Self-supervised VERDICT-MRI for enhanced prostate tumor characterization},
journal = {Magnetic Resonance in Medicine},
volume = {92},
number = {5},
pages = {2181-2192},
url = {https://doi.org/10.1002/mrm.30186},
year = {2024}
}

@inproceedings{kerkela2026,
  title={Set deep learning for protocol generalisation in machine-learning-based brain microstructure estimation},
  author={Kerkel\"a, Leevi and Legouhy, Antoine and Kraguljac, Nina and Zhang, Hui},
  booktitle={ISMRM},
  year={2026},
  url={https://zenodo.org/records/18215491/files/Abstract%20%2301739-7.pdf},
}

@inproceedings{tortoise,
  title={TORTOISE v3: Improvements and new features of the NIH diffusion MRI processing pipeline},
  author={Irfanoglu, Mustafa Okan and Nayak, Amritha and Jenkins, Jeffrey and Pierpaoli, Carlo},
  booktitle={ISMRM 25th annual meeting},
  year={2017}
}

@article{ozarslan2013,
title = {Mean apparent propagator (MAP) MRI: A novel diffusion imaging method for mapping tissue microstructure},
journal = {NeuroImage},
volume = {78},
pages = {16-32},
year = {2013},
issn = {1053-8119},
url = {https://doi.org/10.1016/j.neuroimage.2013.04.016},
author = {Evren Özarslan and Cheng Guan Koay and Timothy M. Shepherd and Michal E. Komlosh and M. Okan İrfanoğlu and Carlo Pierpaoli and Peter J. Basser},
}

@article{bastin1999,
title = {Correction of eddy current-induced artefacts in diffusion tensor imaging using iterative cross-correlation},
journal = {Magnetic Resonance Imaging},
volume = {17},
number = {7},
pages = {1011-1024},
year = {1999},
issn = {0730-725X},
url = {https://doi.org/10.1016/S0730-725X(99)00026-0},
author = {Mark E. Bastin},
}

@article{mangin2002,
title = {Distortion correction and robust tensor estimation for MR diffusion imaging},
journal = {Medical Image Analysis},
volume = {6},
number = {3},
pages = {191-198},
year = {2002},
note = {Special Issue on Medical Image Computing and Computer-Assisted Intervention - MICCAI 2001},
issn = {1361-8415},
url = {https://doi.org/10.1016/S1361-8415(02)00079-8},
author = {J.-F. Mangin and C. Poupon and C. Clark and D. {Le Bihan} and I. Bloch},
}

@article{reese2003,
  title={Reduction of eddy-current-induced distortion in diffusion MRI using a twice-refocused spin echo},
  author={Reese, Timothy G and Heid, Oliver and Weisskoff, Robert M and Wedeen, Van J},
  journal={Magnetic Resonance in Medicine: An Official Journal of the International Society for Magnetic Resonance in Medicine},
  volume={49},
  number={1},
  pages={177--182},
  year={2003},
  publisher={Wiley Online Library}
}

@article{mansfield1986,
title = {Active magnetic screening of gradient coils in NMR imaging},
journal = {Journal of Magnetic Resonance (1969)},
volume = {66},
number = {3},
pages = {573-576},
year = {1986},
issn = {0022-2364},
url = {https://doi.org/10.1016/0022-2364(86)90205-2},
author = {P Mansfield and B Chapman}}

@article{alexander2019,
author = {Alexander, Daniel C. and Dyrby, Tim B. and Nilsson, Markus and Zhang, Hui},
title = {Imaging brain microstructure with diffusion MRI: practicality and applications},
journal = {NMR in Biomedicine},
volume = {32},
number = {4},
pages = {e3841},
doi = {https://doi.org/10.1002/nbm.3841},
url = {https://analyticalsciencejournals.onlinelibrary.wiley.com/doi/abs/10.1002/nbm.3841},
eprint = {https://analyticalsciencejournals.onlinelibrary.wiley.com/doi/pdf/10.1002/nbm.3841},
note = {e3841 NBM-16-0261.R1},
year = {2019}
}

@article{lebihan2012,
  title={Diffusion MRI at 25: exploring brain tissue structure and function},
  author={Le Bihan, Denis and Johansen-Berg, Heidi},
  journal={Neuroimage},
  volume={61},
  number={2},
  pages={324--341},
  year={2012},
  publisher={Elsevier}
}

@article{basser1994,
  title={MR diffusion tensor spectroscopy and imaging},
  author={Basser, Peter J and Mattiello, James and LeBihan, Denis},
  journal={Biophysical journal},
  volume={66},
  number={1},
  pages={259--267},
  year={1994},
  publisher={Elsevier}
}

@article{berman2009,
  title={Diffusion MR tractography as a tool for surgical planning},
  author={Berman, Jeffrey},
  journal={Magnetic resonance imaging clinics of North America},
  volume={17},
  number={2},
  pages={205--214},
  year={2009},
  publisher={Elsevier}
}

@article{fiebach2002,
  title={CT and diffusion-weighted MR imaging in randomized order: diffusion-weighted imaging results in higher accuracy and lower interrater variability in the diagnosis of hyperacute ischemic stroke},
  author={Fiebach, JB and Schellinger, PD and Jansen, O and Meyer, M and Wilde, P and Bender, J and Schramm, P and Juttler, E and Oehler, J and Hartmann, M and others},
  journal={Stroke},
  volume={33},
  number={9},
  pages={2206--2210},
  year={2002},
  publisher={Lippincott Williams \& Wilkins}
}

@article{maier2010,
  title={Diffusion imaging of brain tumors},
  author={Maier, Stephan E and Sun, Yanping and Mulkern, Robert V},
  journal={NMR in biomedicine},
  volume={23},
  number={7},
  pages={849--864},
  year={2010},
  publisher={Wiley Online Library}
}

@article{skare2010,
  title={EPI-based pulse sequences for diffusion tensor MRI},
  author={Skare, ST and Bammer, Roland},
  journal={Diffusion MRI: Theory, Methods, and Applications},
  volume={1},
  pages={182--202},
  year={2011},
  publisher={Oxford University Press Oxford}
}

@article{tortoisev4,
  title={TORTOISEV4: ReImagining the NIH Diffusion MRI Processing Pipeline},
  author={Irfanoglu, M Okan and Nayak, Amritha and Taylor, Paul and Thai, Anh and Pierpaoli, Carlo},
  journal={Imaging Neuroscience},
  year={2025},
  publisher={MIT Press 255 Main Street, 9th Floor, Cambridge, Massachusetts 02142, USA~…},
  url={https://doi.org/10.1162/imag.a.948}
}

@article{chang1992,
  title={A technique for accurate magnetic resonance imaging in the presence of field inhomogeneities},
  author={Chang, Hsuan and Fitzpatrick, J Michael},
  journal={IEEE transactions on medical imaging},
  volume={11},
  number={3},
  pages={319--329},
  year={1992},
  publisher={IEEE}
}

@article{jezzard1998,
  title={Characterization of and correction for eddy current artifacts in echo planar diffusion imaging},
  author={Jezzard, Peter and Barnett, Alan S and Pierpaoli, Carlo},
  journal={Magnetic resonance in medicine},
  volume={39},
  number={5},
  pages={801--812},
  year={1998},
  publisher={Wiley Online Library}
}

@article{andersson2016,
  title={An integrated approach to correction for off-resonance effects and subject movement in diffusion MR imaging},
  author={Andersson, Jesper LR and Sotiropoulos, Stamatios N},
  journal={Neuroimage},
  volume={125},
  pages={1063--1078},
  year={2016},
  publisher={Elsevier}
}

@article{rohde2004,
  title={Comprehensive approach for correction of motion and distortion in diffusion-weighted MRI},
  author={Rohde, Gustavo Kunde and Barnett, AS and Basser, PJ and Marenco, S and Pierpaoli, C},
  journal={Magnetic Resonance in Medicine: An Official Journal of the International Society for Magnetic Resonance in Medicine},
  volume={51},
  number={1},
  pages={103--114},
  year={2004},
  publisher={Wiley Online Library}
}

@inproceedings{isola2017,
  title={Image-to-image translation with conditional adversarial networks},
  author={Isola, Phillip and Zhu, Jun-Yan and Zhou, Tinghui and Efros, Alexei A},
  booktitle={Proceedings of the IEEE conference on computer vision and pattern recognition},
  pages={1125--1134},
  year={2017}
}

@inproceedings{zhu2017,
  title={Unpaired image-to-image translation using cycle-consistent adversarial networks},
  author={Zhu, Jun-Yan and Park, Taesung and Isola, Phillip and Efros, Alexei A},
  booktitle={Proceedings of the IEEE international conference on computer vision},
  pages={2223--2232},
  year={2017}
}

@article{hoopes2022,
  title={SynthStrip: skull-stripping for any brain image},
  author={Hoopes, Andrew and Mora, Jocelyn S and Dalca, Adrian V and Fischl, Bruce and Hoffmann, Malte},
  journal={NeuroImage},
  volume={260},
  pages={119474},
  year={2022},
  publisher={Elsevier}
}

@article{schilling2019,
  title={Synthesized b0 for diffusion distortion correction (Synb0-DisCo)},
  author={Schilling, Kurt G and Blaber, Justin and Huo, Yuankai and Newton, Allen and Hansen, Colin and Nath, Vishwesh and Shafer, Andrea T and Williams, Owen and Resnick, Susan M and Rogers, Baxter and others},
  journal={Magnetic resonance imaging},
  volume={64},
  pages={62--70},
  year={2019},
  publisher={Elsevier}
}

@article{voxelmorph1,
  title={Voxelmorph: a learning framework for deformable medical image registration},
  author={Balakrishnan, Guha and Zhao, Amy and Sabuncu, Mert R and Guttag, John and Dalca, Adrian V},
  journal={IEEE transactions on medical imaging},
  volume={38},
  number={8},
  pages={1788--1800},
  year={2019},
  publisher={IEEE}
}

@inproceedings{devos2017,
  title={End-to-end unsupervised deformable image registration with a convolutional neural network},
  author={De Vos, Bob D and Berendsen, Floris F and Viergever, Max A and Staring, Marius and I{\v{s}}gum, Ivana},
  booktitle={International Workshop on Deep Learning in Medical Image Analysis},
  pages={204--212},
  year={2017},
  organization={Springer}
}

@article{devos2019,
  title={A deep learning framework for unsupervised affine and deformable image registration},
  author={De Vos, Bob D and Berendsen, Floris F and Viergever, Max A and Sokooti, Hessam and Staring, Marius and I{\v{s}}gum, Ivana},
  journal={Medical image analysis},
  volume={52},
  pages={128--143},
  year={2019},
  publisher={Elsevier}
}

@article{andersson2015,
  title={Non-parametric representation and prediction of single-and multi-shell diffusion-weighted MRI data using Gaussian processes},
  author={Andersson, Jesper LR and Sotiropoulos, Stamatios N},
  journal={Neuroimage},
  volume={122},
  pages={166--176},
  year={2015},
  publisher={Elsevier}
}

@inproceedings{unet,
  title={U-net: Convolutional networks for biomedical image segmentation},
  author={Ronneberger, Olaf and Fischer, Philipp and Brox, Thomas},
  booktitle={International Conference on Medical image computing and computer-assisted intervention},
  pages={234--241},
  year={2015},
  organization={Springer}
}

@article{jenkinson2012,
  title={FSL},
  author={Jenkinson, Mark and Beckmann, Christian F and Behrens, Timothy EJ and Woolrich, Mark W and Smith, Stephen M},
  journal={Neuroimage},
  volume={62},
  number={2},
  pages={782--790},
  year={2012},
  publisher={Elsevier}
}

@article{tustison2010,
  title={N4ITK: improved N3 bias correction},
  author={Tustison, Nicholas J and Avants, Brian B and Cook, Philip A and Zheng, Yuanjie and Egan, Alexander and Yushkevich, Paul A and Gee, James C},
  journal={IEEE transactions on medical imaging},
  volume={29},
  number={6},
  pages={1310--1320},
  year={2010},
  publisher={IEEE}
}

@article{billot2023,
	title = {SynthSeg: {Segmentation} of brain {MRI} scans of any contrast and resolution without retraining},
	author = {Billot, Benjamin and Greve, Douglas N. and Puonti, Oula and Thielscher, Axel and Van Leemput, Koen and Fischl, Bruce and Dalca, Adrian V. and Iglesias, Juan Eugenio},
	journal = {{Medical} {Image} {Analysis}},
	year = {2023},
    volume = {86},
	pages = {102789},
    issn = {1361-8415},
    doi = {10.1016/j.media.2023.102789},
}

@article{turner1990,
  title={Echo-planar imaging of intravoxel incoherent motion.},
  author={Turner, Robert and Le Bihan, Denis and Maier, Joseph and Vavrek, Robert and Hedges, L Kyle and Pekar, James},
  journal={Radiology},
  volume={177},
  number={2},
  pages={407--414},
  year={1990}
}

@book{bishop2006,
  title={Pattern recognition and machine learning},
  author={Bishop, Christopher M and Nasrabadi, Nasser M},
  volume={4},
  number={4},
  year={2006},
  publisher={Springer}
}

@article{kraguljac2023,
  title={Neurite orientation dispersion and density imaging in psychiatric disorders: a systematic literature review and a technical note},
  author={Kraguljac, Nina Vanessa and Guerreri, Michele and Strickland, Molly Jordan and Zhang, Hui},
  journal={Biological Psychiatry Global Open Science},
  volume={3},
  number={1},
  pages={10--21},
  year={2023},
  publisher={Elsevier}
}

@article{duong2020,
  title={An unsupervised deep learning technique for susceptibility artifact correction in reversed phase-encoding EPI images},
  author={Duong, Soan TM and Phung, Son L and Bouzerdoum, Abdesselam and Schira, Mark M},
  journal={Magnetic Resonance Imaging},
  volume={71},
  pages={1--10},
  year={2020},
  publisher={Elsevier}
}

@inproceedings{legouhy2022,
  title={Correction of susceptibility distortion in EPI: a semi-supervised approach with deep learning},
  author={Legouhy, Antoine and Graham, Mark and Guerreri, Michele and Stee, Whitney and Villemonteix, Thomas and Peigneux, Philippe and Zhang, Hui},
  booktitle={International Workshop on Computational Diffusion MRI},
  pages={38--49},
  year={2022},
  organization={Springer}
}

@article{epstein2024,
    title = "Choice of training label matters: how to best use deep learning for quantitative MRI parameter estimation",
    author = "Epstein, Sean C. and Bray, Timothy J. P. and Hall-Craggs, Margaret and Zhang, Hui",
    journal = "Machine Learning for Biomedical Imaging",
    volume = "2",
    issue = "January 2024 issue",
    year = "2024",
    pages = "586--610",
    issn = "2766-905X",
    url = "https://doi.org/10.59275/j.melba.2024-geb5"
}

@article{Zhang2012,
    title = {{NODDI: Practical in vivo neurite orientation dispersion and density imaging of the human brain}},
    year = {2012},
    journal = {NeuroImage},
    author = {Zhang, Hui and Schneider, Torben and Wheeler-Kingshott, Claudia A. and Alexander, Daniel C.},
    number = {4},
    pages = {1000--1016},
    volume = {61},
    publisher = {Elsevier Inc.},
    url = {http://dx.doi.org/10.1016/j.neuroimage.2012.03.072},
    doi = {10.1016/j.neuroimage.2012.03.072},
    issn = {10538119},
    pmid = {22484410}
}

@article{Kamiya2020,
    title = {{NODDI in clinical research}},
    year = {2020},
    journal = {Journal of Neuroscience Methods},
    author = {Kamiya, Kouhei and Hori, Masaaki and Aoki, Shigeki},
    number = {August},
    month = {12},
    pages = {108908},
    volume = {346},
    publisher = {Elsevier},
    url = {https://doi.org/10.1016/j.jneumeth.2020.108908 https://linkinghub.elsevier.com/retrieve/pii/S0165027020303319},
    doi = {10.1016/j.jneumeth.2020.108908},
    issn = {01650270},
    pmid = {32814118}
}

@article{andersson2016b,
  title={Incorporating outlier detection and replacement into a non-parametric framework for movement and distortion correction of diffusion MR images},
  author={Andersson, Jesper LR and Graham, Mark S and Zsoldos, Enik{\H{o}} and Sotiropoulos, Stamatios N},
  journal={Neuroimage},
  volume={141},
  pages={556--572},
  year={2016},
  publisher={Elsevier}
}

@article{andersson2017,
  title={Towards a comprehensive framework for movement and distortion correction of diffusion MR images: Within volume movement},
  author={Andersson, Jesper LR and Graham, Mark S and Drobnjak, Ivana and Zhang, Hui and Filippini, Nicola and Bastiani, Matteo},
  journal={Neuroimage},
  volume={152},
  pages={450--466},
  year={2017},
  publisher={Elsevier}
}

\appendix

\section{Details on how to build the transformations from their parameters}
\label{appendix}

This appendix details how the 16 scalar parameters output by the registration model presented in Section~\ref{architecture} are assembled into the
transformation $E$ for the eddy distortions and $R$ for the head motion, following the transformation models defined in Section~\ref{distomodel}. Let $\boldsymbol{c} \in \mathbb{R}^3$ denote the image centre; both transformations 
are expressed relative to $\boldsymbol{c}$ so that they act around the image centre rather than the voxel-grid origin, improving numerical 
conditioning. Let $\boldsymbol{p}\in\mathbb{R}^3$ with $\|\boldsymbol{p}\|=1$ be the vector defining the phase-encoding direction (PED).

\textbf{Rigid transformation.}
The rigid component is parameterised by translation coefficients  $(r_1,r_2,r_3)$ and rotation coefficients $(r_4,r_5,r_6)$. 
The rotation matrix is obtained via the matrix exponential mapping the Lie algebra $\mathfrak{so}(3)$ of skew-symmetric matrices to the Lie group $SO(3)$ of rotation matrices \citep{blanco2021}, which guarantees a valid rotation for any unconstrained network output:
\begin{equation}
\boldsymbol{R}_\mathrm{trans}= (r_1\ r_2\ r_3)^\top, \qquad
\boldsymbol{R}_\mathrm{rot}= \exp\begin{pmatrix} 0 & -r_6 & r_5\\ 
r_6 & 0 & -r_4\\ -r_5 & r_4 & 0 \end{pmatrix}
\end{equation}
where $\boldsymbol{R}_\mathrm{trans}\in \mathbb{R}^3$ and $\boldsymbol{R}_\mathrm{rot}\in SO(3)$. The centred rigid transformation applied to $\boldsymbol{x}$ is:
\begin{equation}
R(\boldsymbol{x})=\boldsymbol{R}_\mathrm{rot}(\boldsymbol{x}-\boldsymbol{c})+\boldsymbol{c}+\boldsymbol{R}_\mathrm{trans}
\end{equation}

\textbf{Eddy transformation}.
The eddy-current component is parameterised by translation coefficient $e_1$, linear coefficients $(e_2,e_3,e_4)$, and quadratic coefficients $(e_5,\dots,e_{10})$, directly corresponding to the $t$, $\boldsymbol{L}$ and $\boldsymbol{Q}$ terms of the distortion model in Section~\ref{distomodel}:
\begin{equation}
E_\mathrm{trans}=e_1,  \qquad
\boldsymbol{E}_\mathrm{lin}=(e_2\ e_3\ e_4), \qquad
\boldsymbol{E}_\mathrm{quad}=\begin{pmatrix}e_5 & e_6 & e_7\\ e_6 & e_8 & e_9\\e_7 & e_9 & e_{10}\end{pmatrix}
\end{equation}
where $E_\mathrm{trans}\in \mathbb{R}$, 
$\boldsymbol{E}_\mathrm{lin}\in \mathbb{R}^{1\times 3}$ and 
$\boldsymbol{E}_\mathrm{quad}\in \mathbb{R}^{3\times 3}_\mathrm{sym}$. 
The eddy transformation applied to $\boldsymbol{x}$ is:
\begin{equation}
E(\boldsymbol{x})= \boldsymbol{x} + \left((\boldsymbol{x}-\boldsymbol{c})^\top \boldsymbol{E}_\mathrm{quad}(\boldsymbol{x}-\boldsymbol{c})
+\boldsymbol{E}_\mathrm{lin}(\boldsymbol{x}-\boldsymbol{c})+E_\mathrm{trans}\right)\boldsymbol{p}
\end{equation}
\begin{remark}
Both transformations are expressed relative to the image centre $\boldsymbol{c}$, so that a zero rotation does not introduce a spurious translation and the polynomial eddy terms are evaluated over centred coordinates of smaller magnitude, improving numerical conditioning. When all 16 parameters are zero, both $R$ and $E$ reduce to the identity, providing a neutral initialisation from which the network can learn corrections.
\end{remark}

\end{document}